\RequirePackage{fix-cm}
\documentclass{article}

\usepackage[table]{xcolor}

\PassOptionsToPackage{hyperfootnotes=false}{hyperref}
\usepackage[final]{corl_2026} 

\usepackage{microtype}
\usepackage{graphicx}
\usepackage{subcaption}
\usepackage{booktabs} 
\usepackage{multirow} 
\usepackage[linesnumbered,ruled,vlined,algo2e]{algorithm2e}
\usepackage[frozencache,cachedir=.]{minted}
\usepackage[most]{tcolorbox}
\SetKwComment{Comment}{$\triangleright$\ }{}

\hypersetup{
    colorlinks,
    pdftitle={Recover, Discover, Plan: Learning Skills and Concepts from Robot Failures},
    pdfauthor={Bowen Li, Mayank Mishra, Y. Isabel Liu, Stone Tao, Nishanth Kumar, Alexander Gray, Ruwan Wickramarachchi, Jonathan Francis, Sebastian Scherer, Tom Silver},
    linkcolor={red!50!black},
    citecolor={blue!50!black},
    urlcolor={blue!80!black}
}

\usepackage{enumitem}

\newtcolorbox{pikebox}[1]{
    enhanced,
    breakable,
    colback=cyan!5!white,
    colframe=blue!50!black,
    fonttitle=\bfseries,
    title={#1},
    width=\linewidth,
    boxrule=0.5pt,
    arc=1mm,
    left=1mm,
    right=1mm,
    top=1mm,
    bottom=1mm
}

\usepackage{amsmath}
\usepackage{amssymb}
\usepackage{mathtools}
\usepackage{amsthm}
\usepackage{pifont}
\newcommand{\cmark}{\ding{51}}
\newcommand{\xmark}{\ding{55}}

\usepackage[capitalize,noabbrev]{cleveref}

\newenvironment{tightlist}%
{\begin{list}{$\bullet$}{%
    \setlength{\topsep}{0in}
    \setlength{\partopsep}{0in}
    \setlength{\itemsep}{0in}
    \setlength{\parsep}{0in}
    \setlength{\leftmargin}{1.5em}
    \setlength{\rightmargin}{0in}
}
}%
{\end{list}
}

\theoremstyle{plain}

\theoremstyle{definition}

\theoremstyle{remark}

\usepackage{tabularx}
\usepackage{wrapfig}

\newcounter{RNum}

\newcommand{\fref}[1]{Figure~\ref{#1}}
\newcommand{\sref}[1]{Section~\ref{#1}}
\newcommand{\tref}[1]{Table~\ref{#1}}
\newcommand{\appref}[1]{Appendix~\ref{#1}}
\newcommand{\algoref}[1]{Algorithm~\ref{#1}}

\newcommand{\myparagraph}[1]{\noindent\textbf{#1}~}

\newcommand{\tablestyle}[2]{\setlength{\tabcolsep}{#1}\renewcommand{\arraystretch}{#2}\centering\footnotesize}

\usepackage[disable,textsize=tiny]{todonotes}

\def\model{ReSYNC}

\title{Recover, Discover, Plan: Learning Skills and Concepts from Robot Failures}

\author{
Bowen Li$^{1,5}$\thanks{Correspondence to: \texttt{bowenli2@andrew.cmu.edu}, \texttt{basti@andrew.cmu.edu},
and \texttt{tsilver@princeton.edu}, work partly done when Bowen Li was an intern at Centaur AI.}~,
Mayank Mishra$^{1}$,
Y.\ Isabel Liu$^{2}$,
Stone Tao$^{3}$,
Nishanth Kumar$^{4}$,
\\
\textbf{
Alexander Gray$^{5}$,
Ruwan Wickramarachchi$^{6}$,
Jonathan Francis$^{1,6}$,
}
\\
\textbf{
Sebastian Scherer$^{1}$,
Tom Silver$^{2}$
}
\\
$^1$CMU, $^2$Princeton, $^3$UCSD, $^4$MIT, $^5$Centaur AI, $^6$Bosch Center for AI
}

\makeatletter
\renewcommand{\@conferencelocation}{Austin, Texas, USA}
\makeatother

\begin{document}
\maketitle


{\centering \url{https://jaraxxus-me.github.io/ReSYNC/}\par}

\begin{abstract}
Intelligent robots should not only recover from failures, but also acquire the abstract knowledge needed to avoid them in the future.
While reinforcement learning (RL) can learn reactive recovery behaviors, training a separate policy for every distinct failure mode is highly inefficient.
We introduce Recovery-Driven Synthesis of Relational Concepts (\model{}), the first approach that progressively discovers and refines state abstractions (relational predicates) from failure-recovery experience to support abstract planning.
Unlike purely reactive methods, \model{} jointly learns skills and concepts through an incremental dual-learning process.
In the skill-learning phase, the robot uses RL to learn to recover from failures seen in training tasks.
In the concept-learning phase, the robot discovers new relational predicates and refines its abstract planning model to explain and generalize the learned recovery behaviors.
This interaction enables \model{} to convert local recoveries seen during training into global failure avoidance at test time.
Across four simulated domains, we show that \model{}'s ability to continually expand and refine its abstraction library allows it to solve long-horizon, previously unseen problems, outperforming strong baselines by over 50\%. 
Additionally, we demonstrate sim-to-real transfer of \model{}, where it performs real-world non-prehensile manipulation skills and generalizes to unseen scenarios through abstract planning.
Overall, \model{} represents a significant step toward robots that autonomously acquire abstractions for scalable, failure-aware planning in the physical world.
\end{abstract}




\section{Introduction}

Failure is inevitable for a robot in a \emph{big world}~\cite{javed2024big}.
A rover may slip on ice or get trapped in mud.
A mobile manipulator may struggle to retrieve an object from a tight drawer.
Some failures can be anticipated and avoided by design, but truly intelligent robots should handle failures on their own.

Previous work has proposed that robots should learn \emph{recovery skills} to escape from failures~\cite{vats2023recovery,vats2024recoverychaining,bagaria2021dsg,bagaria2025im-dsg,li2024bridge}.
Recovery skills perform best when some failure modes are repeatedly encountered.
But when the world is \emph{big} and novel failure configurations are combinatorial, recovery skill learning alone scales poorly. It treats failures as isolated events rather than exploiting their shared structure.

\begin{figure}[!t]
	\centering
    \includegraphics[width=1\columnwidth]{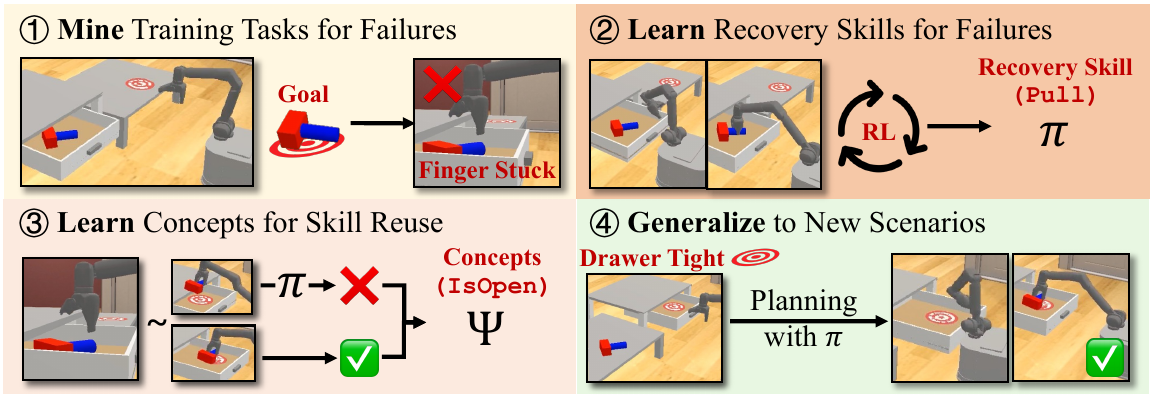}
    \caption{Overview of \model{}.
    The learning in \model{} is driven by failures in training tasks, where it starts by acquiring a recovery skill from environmental interaction (opening a drawer before picking).
    Then, \model{} discovers abstract concepts that ground recovery skills in a new abstract planning model.
    At test time, \model{} uses abstract planning to anticipate and avoid failures.}
    \vspace{-0.3cm}
	\label{fig:teaser}
\end{figure}

We propose that robots should treat failures not as impediments necessitating recovery, but rather, as opportunities for \emph{discovery}.
We consider a robot with (1) a general library of skills and concepts that are grounded in the physical world; (2) a symbolic planner that composes the skills and concepts given a task; and (3) a curriculum of training tasks designed to elicit failures.
When the robot encounters a failure, it initiates a skill-concept dual learning process that augments the skills and concepts in its library.
As illustrated in \fref{fig:teaser}, the robot augments its skill set by learning a new recovery skill (pull the drawer) to resolve a failed $\mathtt{Pick}$, and revises its planning abstractions by discovering new concepts such as $\mathtt{IsOpen}$ that modify the preconditions of existing skills.
After learning, the robot not only recovers from the same failure if it occurs again, but also generalizes to a combinatorial number of unseen failures by recomposing the new skills and concepts that it has discovered.

From the perspective of abstract planning, prior work typically assumes hand-designed skills for learning concepts~\cite{silver2023predicateinvent,shah2024reals,li2025IVNTR}, hand-designed concepts for learning skills~\cite{silver2022skills,liu2025slap}, or substantial supervision when learning both~\cite{shao2025symskill}. 
\model{} relaxes these assumptions by using failure-recovery \textit{experience}: failures expose missing skills, while recoveries generate self-supervised data for discovering concepts that make these skills reusable.
From the perspective of reinforcement learning (RL)~\cite{abel2018state,eysenbach2018diversity,bagaria2025im-dsg,nayyar2025autonomous}, \model{} is a model-based, hierarchical approach that leverages symbolic structure and significant inductive bias to solve a broad distribution of long-horizon sparse-reward tasks efficiently.
In both cases, our key idea is to \emph{use recovery to drive progressive skill and concept discovery}.

We evaluate \model{} across four physical environments with continuous state and action spaces~\cite{shuklamaniskill}. 
Our results show that \model{} achieves an average success rate of $70\%$ on compositionally novel tasks, outperforming state-of-the-art recovery (and predicate) learning~\cite{vats2024recoverychaining,li2025IVNTR}, policy learning~\cite{battaglia2018gnn,vaswani2017tf}, VLM planning~\cite{openai2026gpt54thinking}, and hierarchical reinforcement learning (HRL) baselines~\cite{bagaria2025im-dsg}.
We further demonstrate sim-to-real transfer of \model{}, where it learns a non-prehensile manipulation skill for recovery and generalizes to new scenarios via discovered concepts.
Overall, these results suggest that recovery-driven skill and concept discovery is a promising approach for scaling robots to big worlds.

\section{Related Work}

\myparagraph{Learning Abstractions for Hierarchical Planning:}
Hierarchical planners~\cite{kaelbling2011htn,garrett2020pddlstream} address long-horizon tasks using abstractions such as predicates, skills, and operators to decompose problems into tractable sub-goals. 
Recent work~\cite{chitnis2021nsrt,silver2022skills,kumar2024practice,shao2025symskill,konidaris2018skills,wang2025unipred} studies learning these abstractions from data to reduce manual engineering. 
\model{} is particularly motivated by recent effect-centric predicate discovery frameworks~\cite{li2025IVNTR,wang2025unipred}, which learn relational predicates grounded by neural networks while assuming a fixed skill library and an offline demonstration dataset. 
In contrast, \model{} jointly synthesizes both skills and concepts directly from task-driven interactions without explicit supervision.

\myparagraph{Hierarchical Reinforcement Learning:}
HRL commonly adopts the options framework~\cite{sutton1999between} to discover temporal abstractions and optimize high-level policies. 
Representative approaches include intrinsic skill discovery~\cite{eysenbach2018diversity,bagaria2025im-dsg}, option-critic methods~\cite{bacon2017oc,zhang2019dac}, and contrastive skill learning~\cite{moon2023discovering,eysenbach2022contrastive}. 
Deep Skill Graphs (DSG)~\cite{bagaria2021dsg,bagaria2025im-dsg} is particularly related, as it expands a planning graph by discovering new low-level skills. 
Unlike DSG, \model{} learns relational predicates that optimize an explicit abstract world model, leading to improved planning and sample efficiency.

\myparagraph{Integrated RL and Planning:}
A growing body of work combines reinforcement learning with symbolic planning. 
Prior methods construct reachability graphs from replay buffers~\cite{eysenbach2019search,savinov2018semi}, learn RL policies to guide planning~\cite{yang2018peorl}, or discover new skills on planning graphs~\cite{liu2025slap}. 
More closely related are recovery learning methods~\cite{jiang2018integrating,li2024bridge,vats2024recoverychaining,sarathy2020spotter,goel2022rapid}, which use RL to recover from states where planning fails under novel conditions. 
However, existing approaches are largely reactive: recovery skills are triggered only after failure and do not modify the underlying abstractions. 
In contrast, \model{} discovers new concepts that ground the recovery experience, enabling skill reuse in unseen tasks.

\section{Background and Problem Setting}\label{sec:bg}
We mainly follow the notation system from previous work~\cite{li2025IVNTR}; see \appref{app:notation} for a glossary.

\myparagraph{Environment Modeling:} We model the environment as a Relational Factored Markov Decision Process (RF-MDP).
The state $\mathbf{x} \in \mathcal{X}$ is factored by typed objects $\mathtt{o}\in\mathcal{O}$ with continuous features $\mathbf{x}_{\mathtt{o}}$ (e.g., 6D pose).
A sparse-reward task $T=\left\langle\mathcal{O}, \mathbf{x}_0, g\right\rangle$ specifies objects, an initial state, and a goal $g \subseteq \mathcal{X}$.
A solution is a sequence of low-level actions $\mathbf{a}\in\mathcal{A}$ that leads to the goal.
During training, the robot interacts with a resettable dynamics model $f(\mathbf{x}'\mid \mathbf{x}, \mathbf{a})$ and has access to a broad initial state distribution $P(\mathbf{x}_0)$; at test time, it must achieve $g$ in one finite-horizon rollout.
The dynamics are stationary across resets but need not be deterministic; contact-rich interactions may be stochastic.

\begin{wrapfigure}{r}{0.508\textwidth}
  \centering
  \vspace{-1.0em}
  \includegraphics[width=0.5\textwidth]{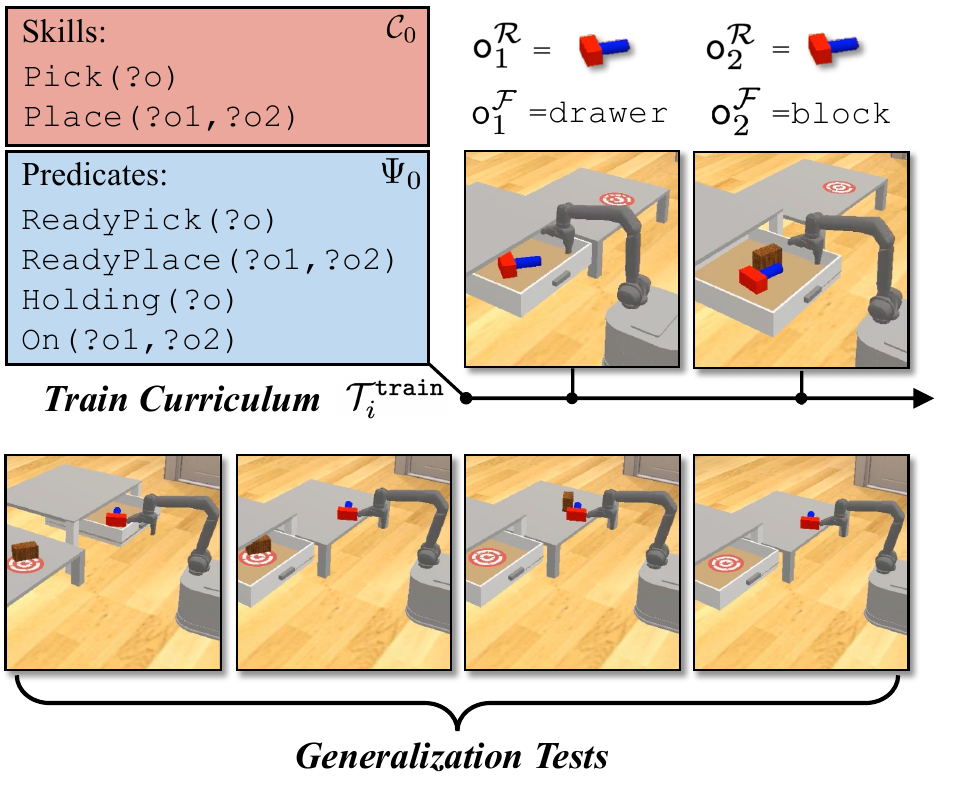}
  \vspace{-0.5em}
  \caption{Overview of the running example (Cluttered Drawer). We show the initial knowledge available to the robot, each stage of the curriculum, and example generalization tests.}
  \label{fig:example}
\end{wrapfigure}

\myparagraph{Planning Abstractions and Execution:}
The robot uses concepts (predicates, $\Psi$) and skills ($\mathcal{C}$) to solve long-horizon tasks with sparse rewards.
A concept is a state predicate. A predicate $\psi \in \Psi$ has a type signature and classifier $\theta_\psi$; a ground predicate $\underline{\psi}$ binds $\psi$ to objects with truth value $\theta_{\underline{\psi}}(\mathbf{x})$.
Each skill is an option model~\cite{sutton1999between}: $\mathtt{C} \in \mathcal{C}$ has a type signature, a policy $\pi_\mathtt{C}$, and a terminal function $\beta_\mathtt{C}$; a ground skill $\underline{\mathtt{C}}$ binds it to objects.
Following prior works~\cite{li2025IVNTR,wang2025unipred}, we associate skills $\mathcal{C}$ with an operator set $\mathtt{Op}^{\mathcal{C}}$ using Planning Domain Definition Language (PDDL)~\cite{McDermott1998PDDL}. Each operator specifies preconditions and add/delete effects over predicates, describing when its skill can execute and the expected symbolic transition. Thus, predicates classify grounded object-centric features, while operators describe transitions between symbolic states; see \appref{app:resync_details}.
Given $\mathbf{x}_0$ and $g$, abstract planning maps them into symbolic representations via $\Psi$, searches for ground skills using $\mathtt{Op}^{\mathcal{C}}$~\cite{helmert2006fast}, executes each skill until termination, and replans when observed effects deviate.
We denote this integrated planning and execution process as $\mathrm{Plan}_f(\mathbf{x} \mid \mathcal{C}, \Psi, \mathtt{Op}^{\mathcal{C}}, g)$, yielding the final state after executing the last skill.

\myparagraph{Initial Knowledge:} The robot starts with initial skills $\mathcal{C}_0$, concepts $\Psi_0$, and operators $\mathtt{Op}^{\mathcal{C}_0}$ that solve simple tasks, such as relocating an unobstructed hammer (\fref{fig:example}).
Constructing this initial library is complementary to our contribution: augmenting skills and concepts from failures.
However, these initial abstractions are often oblivious to complexity; for example, they may implicitly assume an object is always reachable, failing to account for closed drawers or obstructing blocks.

\myparagraph{Training Curriculum:} A user-specified curriculum presents related tasks in stages to elicit failures.
Let $\mathcal{T}_i^{\mathtt{train}}$ be the $i^\text{th}$ stage with $N_i=10$ tasks in our experiments.
We assume the robot can detect failure states (e.g., simulator collisions)~\cite{vats2024recoverychaining} and identify responsible objects through environmental interaction.
For simplicity, we next describe the method assuming a single failure object per stage, denoted $\mathtt{o}_i^{\mathcal{F}}$; in the running example (\fref{fig:example}), $\mathtt{o}_1^{\mathcal{F}}=\mathtt{drawer}$ and $\mathtt{o}_2^{\mathcal{F}}=\mathtt{block}$.
See \sref{sec:experiments} and \appref{app:resync_details} for situations with multiple failure objects.

\myparagraph{Test Time:} After training, the robot is evaluated on held-out tasks with different and more objects than those seen during training.
We measure success as achieving the task goal within a finite horizon.
A naive robot that ignores the training curriculum frequently fails at test time, while merely learning recovery skills and failure detectors is still insufficient for generalization~\cite{vats2024recoverychaining}.
For example, after learning to open a drawer to recover from a failed grasp, the robot should also infer that it must first put down a held object before opening the drawer in a new context.
Thus, beyond skills, the robot must also learn concepts that enable reasoning over skill composition for test-time generalization.
We call recovery \emph{reactive} when it is triggered \textit{only} after failure detection, and failure avoidance \emph{proactive} when the planner invokes a learned skill beforehand to prevent a predicted failure.

\section{Methodology}\label{sec:method}

\begin{figure*}[!t]
	\centering
	\includegraphics[width=\textwidth]{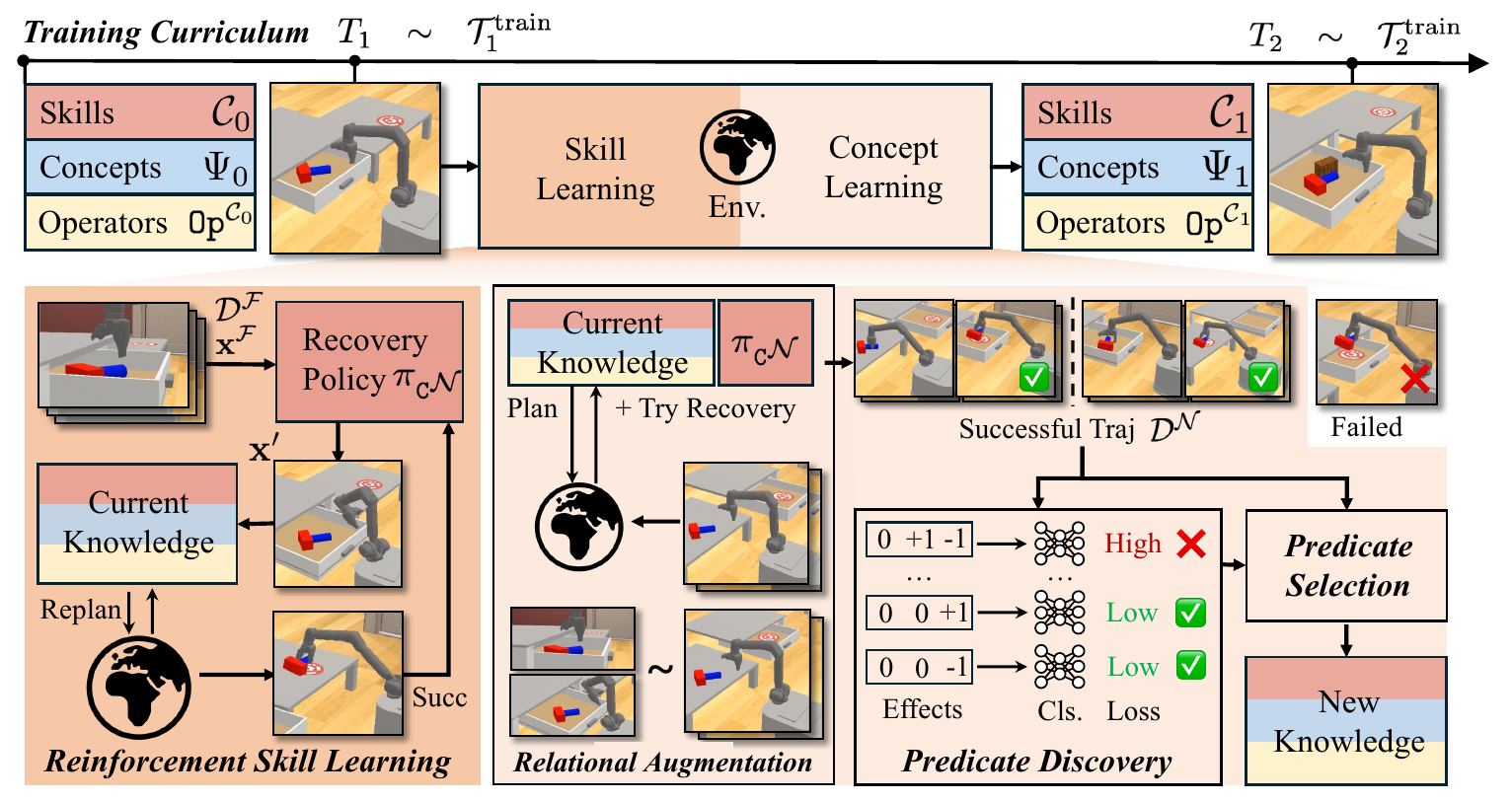}
    \caption{The \model{} framework.
    \textbf{Top:} \model{} alternates between recovery skill learning and concept discovery in each stage.
    \textbf{Bottom left:} Skill learning uses successful replanning as the recovery reward.
    \textbf{Bottom right:} Relational data augmentation supplies trajectories for predicate and operator learning.}
    \vspace{-1em}
	\label{fig:main}

\end{figure*}

We now present \model{}, our method for recovery-driven skill and concept synthesis (Figure~\ref{fig:main}).
\myparagraph{Overview:}
\model{} learns from a staged curriculum by progressively augmenting its skill and concept library.
At each stage, the robot plans until failure, mines related failures, and learns an RL recovery skill that reaches states from which planning can resume (Section~\ref{sec:skill_learning}).
It then uses skill-generated data to discover predicates and update operators, making the new skill composable by the planner (Section~\ref{sec:concept_learning}).
The updated library is reused in later stages (Section~\ref{sec:progressive_learning}) and, at test time, supports planning on unseen tasks.
Unless otherwise specified, we omit stage subscripts below.

\subsection{Skill Learning}\label{sec:skill_learning}
\myparagraph{Failure Mining:}
For each training task $T_j \in \mathcal{T}^{\mathtt{train}}$, \model{} plans and executes with its current abstractions until failure, producing $\mathbf{x}_j^{\mathcal{F}}$.
Within a stage, tasks share the goal $g$, failure object $\mathtt{o}^{\mathcal{F}}$, and failed skill $\mathtt{C}^{\mathcal{F}}$; otherwise, tasks can be split into more stages.
To expand the few failures $\{\mathbf{x}_j^{\mathcal{F}}\}_{j=1}^{N}$, \model{} resamples $\mathtt{o}^{\mathcal{F}}$ relative to a \emph{relevant object} $\mathtt{o}^{\mathcal{R}}$ derived from $\underline{\mathtt{C}}^{\mathcal{F}}$~\cite{silver2022skills}; e.g., for the hammer-picking collision with a block, $\mathtt{o}^{\mathcal{F}}=\mathtt{block}$ and $\mathtt{o}^{\mathcal{R}}=\mathtt{hammer}$.
Using $\{\mathbf{x}_j^{\mathcal{F}}\}_{j=1}^{N}$, \model{} fits pairwise Gaussians over the relative states of $\mathtt{o}^{\mathcal{F}}$ and $\mathtt{o}^{\mathcal{R}}$, resamples $\mathtt{o}^{\mathcal{F}}$ in new initial states $\mathbf{x}_0'\sim P(\mathbf{x}_0)$, and replans.
It keeps only resulting failure states, yielding the augmented dataset $\mathcal{D}^{\mathcal{F}}$.

\myparagraph{Policy and Terminal Learning:}
\model{} next constructs a \emph{recovery learning MDP} from $\mathcal{D}^\mathcal{F}$ and the original task goal $g$.
The MDP shares the same dynamics as the environment, but initializes from states in $\mathcal{D}^\mathcal{F}$ and defines success as reaching a state $\mathbf{x}'$ from which $\mathrm{Plan}_f(\mathbf{x}' \mid \mathcal{C}, \Psi, \mathtt{Op}^{\mathcal{C}}, g)$ succeeds.
Here, $\mathbf{x}'$ is a recovery target (a sub-goal for PPO), rather than necessarily a state satisfying the original task goal $g$.
We use PPO~\cite{schulman2017ppo} to learn a recovery policy $\pi_{\mathtt{C}^\mathcal{N}}$ and train a terminal function $\beta_{\mathtt{C}^\mathcal{N}}$ via supervised learning.
Both the policy and the terminal function are relational~\cite{silver2022skills}: given a state $\mathbf{x}$, we apply a projection $\mathrm{proj}(\mathbf{x})$ containing only features of $\mathtt{o}^{\mathcal{R}}$ and $\mathtt{o}^{\mathcal{F}}$.
The learned skill $\mathtt{C}^\mathcal{N}$ can then be grounded with different objects of the same types.
The library is updated with $\mathtt{C}^\mathcal{N}$, but the planner cannot yet leverage it without a symbolic description.
For that, we proceed to concept learning.

\myparagraph{Efficient PPO Training:}
Following the recovery-learning setting of RC~\cite{vats2024recoverychaining}, PPO executes for a fixed horizon, and recovery success is evaluated only at the terminal state.
Early in training, when successful exploration is rare, we label terminal states by running PDDL replanning and executing the resulting plan: a terminal state is successful if this continuation reaches $g$.
To accelerate these evaluations, we implement skills as batched PyTorch functions and execute a single shared plan across parallel environments.
Once enough successful and failed terminal states have been collected, we train a supervised success classifier and use its predictions as a reward surrogate for subsequent PPO training.
This avoids repeatedly running planning and execution to evaluate recovery success during the remainder of training.
Domain-specific rollout horizons and training budgets are provided in \appref{app:domain_details}, and terminal-function supervision is described in \appref{app:resync_details}.

\subsection{Concept Learning}\label{sec:concept_learning}
A key contribution of \model{} is that it not only recovers from failures observed during training, but also discovers concepts that support recovery skill reuse at test time.
After adding a recovery skill $\mathtt{C}^{\mathcal{N}}$ to form $\mathcal{C}' = \mathcal{C} \cup \{\mathtt{C}^{\mathcal{N}}\}$, \model{} learns predicates $\Psi'$ and operators $\mathtt{Op}^{\mathcal{C}'}$ that enable abstract planning with $\mathtt{C}^{\mathcal{N}}$.
For example, after learning to pull a drawer before grasping a hammer, the robot should discover $\mathtt{IsOpen}$ so the skill can later be reused before placing objects into the drawer.

\myparagraph{Offline Concept Learning with IVNTR:}
To learn these abstractions, \model{} builds on IVNTR~\cite{li2025IVNTR}, a bilevel learning framework for jointly discovering predicate symbols and training their neural classifiers.
IVNTR alternates between symbolic structure learning and neural classifier optimization, then searches over predicate subsets using a planning-efficiency surrogate objective~\cite{silver2023predicateinvent}.
We refer readers to \citet{li2025IVNTR} and \appref{app:resync_details} for additional details.
However, IVNTR assumes access to a large transition dataset $\mathcal{D}^{\mathcal{N}}$ that broadly covers the states encountered during planning and execution.
In our setting, no demonstrations are available and new skills are learned online. \model{} therefore augments the presented recovery data so that IVNTR can be applied for concept learning.

\myparagraph{Relational Data Augmentation:}
A simple data source would be the trajectories collected during recovery skill learning (\sref{sec:skill_learning}).
By definition, this experience includes sequences of skills leading up to the new recovery skill, the recovery skill itself, and a final sequence of skills that completes the training task.
However, such data only captures the original recovery context (e.g., pulling a drawer to grasp a hammer) and does not contain examples of recovery skill reuse (e.g., pulling the drawer for placing another object).
In other words, we have data for $\mathtt{Pull(hammer)}$, but not for
$\mathtt{Pull(target)}$.
The latter data is important for disambiguating $\mathtt{IsOpen}$ from other possible explanations of successful recovery, e.g., $\mathtt{HammerInOpenDrawer}$.

To collect reuse examples, \model{} performs \emph{relational data augmentation}, a self-supervised process that augments recovery experience through simulator interaction.
As in failure mining, relational data augmentation creates counterfactual tasks by resampling $\mathtt{o}^{\mathcal{F}}$ relative to other objects.
We apply sampling to all type-compatible groundings, rather than only the object binding seen during recovery training.
The robot plans in these tasks, invokes the new recovery skill when failure occurs, resumes planning after termination, and adds successful transitions to the offline concept-learning dataset.

Relational data augmentation differs in how it resamples initial states: it targets states \emph{before and after skill reuse}, such as states before and after executing $\mathtt{Pull(target)}$.
Recall that we previously fit pairwise Gaussian distributions to capture the relative states between $\mathtt{o}^{\mathcal{F}}$ and $\mathtt{o}^{\mathcal{R}}$.
These distributions correspond to states \emph{before} the \emph{original} recovery skill (e.g., $\mathtt{Pull(hammer)}$).
To generate data that capture skill reuse on other objects, we additionally fit pairwise Gaussian distributions after the original recovery skill.
We then resample $\mathtt{o}^{\mathcal{F}}$'s state from one of the two pairwise Gaussian distributions, now using additional reference objects beyond the original $\mathtt{o}^{\mathcal{R}}$.
For instance, sampling an open $\mathtt{drawer}$ near a $\mathtt{target}$ creates data where the drawer is open but empty, separating $\mathtt{IsOpen}$ from $\mathtt{HammerInOpenDrawer}$.

\subsection{Progressive Learning}\label{sec:progressive_learning}

After the first stage, \model{} enters later stages with the latest skills, concepts, and operators, progressively expanding its library for increasingly complex unseen tasks.
Two designs support multi-stage concept learning: compositional relational augmentation and concept fine-tuning.

\myparagraph{Compositional Relational Augmentation:}
As training progresses, the relational data augmentation process becomes compositional.
Failure objects $\mathtt{o}^{\mathcal{F}}_v, v \in \{1,\dots,i\}$ from previous stages, together with their learned state distributions, are combinatorially composed with objects matching the types of $\mathtt{o}_v^{\mathcal{R}}$.
This process produces increasingly diverse trajectories that improve learned concepts.

\myparagraph{Concept Fine-tuning:}
A simple strategy to manage the concepts would be relearning them from scratch after each stage using all of the data, but this is computationally inefficient.
Instead, \model{} retains previously learned concepts and incrementally expands the abstraction library using newly generated data.
However, freezing existing concepts can lead to inconsistencies when newly learned skills alter previously observed state distributions.
For example, a recovery skill that pushes a $\mathtt{block}$ away from a $\mathtt{hammer}$ may perturb a nearby $\mathtt{drawer}$, causing the predicate $\mathtt{IsOpen}$ to become inaccurate.
To address this issue, \model{} fine-tunes the classifiers for existing concepts before discovering new ones, ensuring that they remain consistent with the expanded skill set and the newly observed states.
In particular, we extend IVNTR to handle the case where some symbolic concepts are known but others must be discovered (and all classifiers must be trained). See \appref{app:resync_details}.

\section{Experiments}
\label{sec:experiments}

We use experiments to answer the following questions about \model{}'s efficacy and efficiency:
\begin{enumerate}[topsep=2pt, itemsep=1pt, parsep=0pt]
    \item[\textbf{Q1.}] Can \model{} recover from failures and generalize?
    \item[\textbf{Q2.}] Does \model{} progressively learn without forgetting?
    \item[\textbf{Q3.}] Can \model{} be deployed in the real world for planning and recovery?
    \item[\textbf{Q4.}] Do compositional relational augmentation and fine-tuning help?
    \item[\textbf{Q5.}] To what extent does \model{} perform sample-efficient learning during training time?
\end{enumerate}
\vspace{-1em}

\begin{figure*}[!t]
	\centering
	\includegraphics[width=\textwidth]{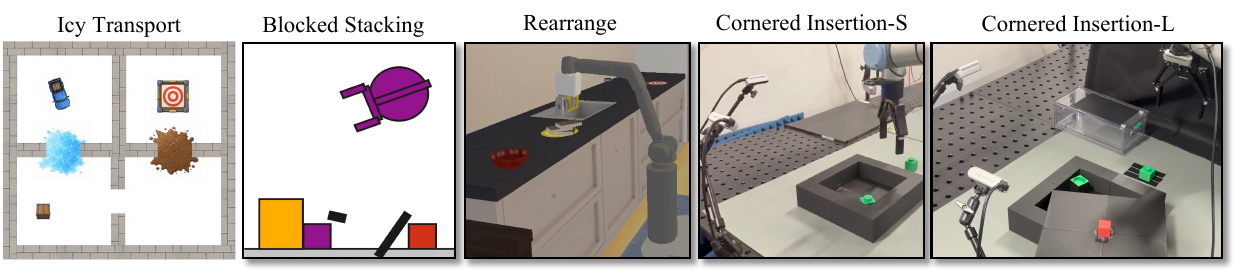}
	\caption{Visualization of the three simulated domains (excluding Cluttered Drawer) and the two real-world domains. Skill videos are available on the project website.}
	\label{fig:domains}
    \vspace{-1em}
\end{figure*}

\begin{figure*}[!t]
	\centering
	\includegraphics[width=\textwidth]{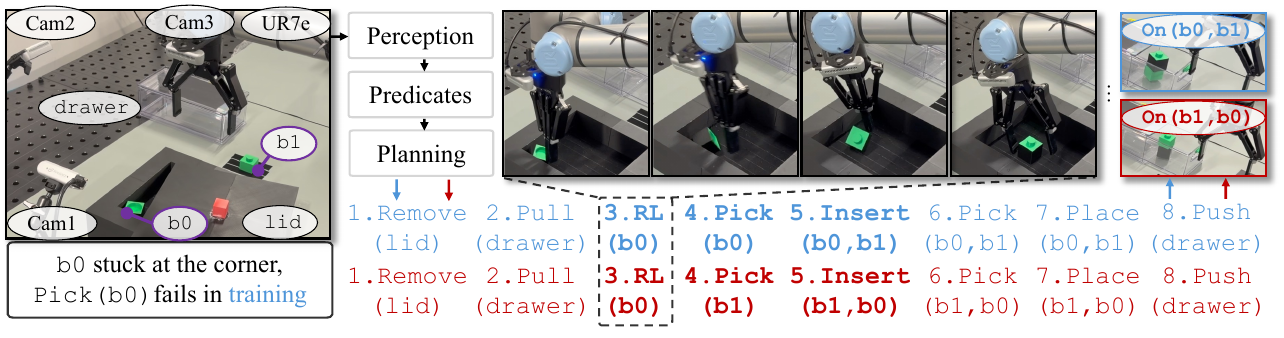}
    \vspace{-1em}
	\caption{Hardware setup and \model{} pipeline in the real-world experiments. During training (blue), the robot fails to pick \texttt{b0} and learns an RL skill for recovery.
    In unseen test tasks with a different goal (red), \model{} plans to first use the RL skill on \texttt{b0} to avoid the insertion failure.}
    \vspace{-1em}
	\label{fig:real}
\end{figure*}

\subsection{Implementation Details}
\myparagraph{Hardware:} Simulated experiments were conducted on an NVIDIA A100 GPU with an AMD EPYC 7543 CPU. For real experiments, models were trained on an NVIDIA RTX PRO 5000 GPU and evaluated on an NVIDIA RTX 3090 GPU.
In \fref{fig:real}, the real setup uses a UR7e arm with a Robotiq 2F-140 gripper and three RealSense D435i cameras providing side, front, and wrist views.

\myparagraph{Baselines:} To ensure fairness, all baselines are given the same initial knowledge and learned recovery skills. We compare \model{} against Recovery Chaining (RC)~\cite{vats2024recoverychaining}, RC+IVNTR~\cite{li2025IVNTR}, Deep Skill Graph single-task (DSG-S) and multi-task (DSG-M)~\cite{bagaria2021dsg,bagaria2025im-dsg}, a VLM planner with one in-context example (VLM-OS, GPT-5.4~\cite{openai2026gpt54thinking}), and supervised GNN-Policy and TF-Policy~\cite{battaglia2018gnn, vaswani2017tf} trained on $\mathcal{D}^\mathcal{N}$. RC retains the initial symbolic planner and invokes recovery after failure. RC+IVNTR and DSG-M incorporate recovery skills into symbolic planning, but use different abstraction-learning procedures. Additional details and analysis are provided in \appref{app:baseline_details}.

\begin{table*}[!t]
    \centering
    \setlength{\tabcolsep}{0.55mm}
    \fontsize{7.5}{9}\selectfont
    \begin{tabular}{c|cc|ccc|ccc|cc}
    \toprule[1.5pt]
    Domain & \multicolumn{5}{c|}{Icy Transport}    & \multicolumn{5}{c}{Cluttered Drawer} \\
    \midrule
    Test Dist. & $\mathcal{T}^\mathrm{train}_1$    & $\mathcal{T}^\mathrm{test}_1$   & $\mathcal{T}^\mathrm{train}_2$    & $\mathcal{T}^\mathrm{test}_2$   & $\mathcal{T}_1$   &$\mathcal{T}^\mathrm{train}_1$    & \multicolumn{1}{c|}{$\mathcal{T}^\mathrm{test}_1$}  & \multicolumn{1}{c}{$\mathcal{T}^\mathrm{train}_2$}    & $\mathcal{T}^\mathrm{test}_2$   & $\mathcal{T}_1$ \\
    Horizon & $3000$ & $3000$ & $5000$ & $5000$ & $5000$ & $500$ & \multicolumn{1}{c|}{$500$} & \multicolumn{1}{c}{$800$} & $800$ & $800$ \\
    \midrule
    \multicolumn{1}{c|}{GNN~\cite{battaglia2018gnn}} & \textbf{0.98}$_{\pm.03}$ & $0.01_{\pm.01}$ & $0.18_{\pm.13}$ & $0.18_{\pm.13}$ & \multicolumn{1}{c|}{$0.38_{\pm.18}$} & $0.86_{\pm.04}$ & \multicolumn{1}{c|}{$0.00_{\pm.00}$} & \multicolumn{1}{c}{\textbf{0.69}$_{\pm.07}$} & $0.48_{\pm.10}$ & $0.00_{\pm.00}$ \\
    \multicolumn{1}{c|}{TF~\cite{vaswani2017tf}} & \textbf{0.98}$_{\pm.03}$ & $0.00_{\pm.00}$ & $0.26_{\pm.14}$ & $0.17_{\pm.12}$ & \multicolumn{1}{c|}{$0.52_{\pm.16}$} & $0.11_{\pm.07}$ & \multicolumn{1}{c|}{$0.00_{\pm.00}$} & \multicolumn{1}{c}{$0.03_{\pm.03}$} & $0.02_{\pm.02}$ & $0.00_{\pm.00}$ \\
    \multicolumn{1}{c|}{VLM-OS~\cite{openai2026gpt54thinking}} & $0.39_{\pm.13}$ & $0.00_{\pm.00}$ & $0.30_{\pm.14}$ & $0.00_{\pm.00}$ & \multicolumn{1}{c|}{$0.24_{\pm.12}$} & $0.68_{\pm.06}$ & \multicolumn{1}{c|}{$0.29_{\pm.03}$} & \multicolumn{1}{c}{$0.18_{\pm.04}$} & $0.25_{\pm.06}$ & $0.50_{\pm.10}$ \\
    \multicolumn{1}{c|}{DSG-S~\cite{bagaria2025im-dsg}} & $0.67_{\pm.12}$ & $0.00_{\pm.00}$ & $0.24_{\pm.07}$ & $0.02_{\pm.02}$ & \multicolumn{1}{c|}{$0.44_{\pm.08}$} & $0.73_{\pm.03}$ & \multicolumn{1}{c|}{$0.04_{\pm.01}$} & \multicolumn{1}{c}{$0.67_{\pm.04}$} & $0.04_{\pm.01}$ & $0.38_{\pm.02}$ \\
    \multicolumn{1}{c|}{DSG-M~\cite{bagaria2025im-dsg}} & $0.61_{\pm.04}$ & $0.33_{\pm.04}$ & $0.15_{\pm.07}$ & $0.01_{\pm.02}$ & \multicolumn{1}{c|}{$0.21_{\pm.05}$} & $0.13_{\pm.02}$ & \multicolumn{1}{c|}{$0.03_{\pm.01}$} & \multicolumn{1}{c}{$0.00_{\pm.00}$} & $0.00_{\pm.00}$ & $0.09_{\pm.04}$ \\
    \multicolumn{1}{c|}{RC~\cite{vats2024recoverychaining}} & $0.71_{\pm.11}$ & $0.00_{\pm.00}$ & $0.34_{\pm.16}$ & $0.00_{\pm.00}$ & \multicolumn{1}{c|}{$0.50_{\pm.18}$} & $0.68_{\pm.03}$ & \multicolumn{1}{c|}{$0.01_{\pm.01}$} & \multicolumn{1}{c}{$0.62_{\pm.04}$} & $0.14_{\pm.02}$ & $0.35_{\pm.15}$ \\
    \multicolumn{1}{c|}{RC+IVNTR~\cite{li2025IVNTR}} & $0.50_{\pm.20}$ & $0.44_{\pm.18}$ & $0.18_{\pm.12}$ & $0.08_{\pm.08}$ & \multicolumn{1}{c|}{$0.66_{\pm.16}$} & $0.69_{\pm.03}$ & \multicolumn{1}{c|}{$0.59_{\pm.15}$} & \multicolumn{1}{c}{$0.64_{\pm.03}$} & $0.27_{\pm.18}$ & $0.58_{\pm.14}$ \\
    \multicolumn{1}{c|}{\textbf{ReSYNC}} & $0.92_{\pm.05}$ & $\textbf{0.76}_{\pm.16}$ & $\textbf{0.85}_{\pm.09}$ & $\textbf{0.68}_{\pm.13}$ & \multicolumn{1}{c|}{$\textbf{0.87}_{\pm.09}$} & $\textbf{0.95}_{\pm.01}$ & \multicolumn{1}{c|}{$\textbf{0.96}_{\pm.02}$} & \multicolumn{1}{c}{$0.64_{\pm.02}$} & $\textbf{0.72}_{\pm.03}$ & $\textbf{0.76}_{\pm.03}$ \\
    \midrule
    \midrule
    Domain & \multicolumn{8}{c|}{Blocked Stacking}                          & \multicolumn{2}{c}{Rearrange} \\
    \midrule
    Test Dist. & $\mathcal{T}^\mathrm{train}_1$    & $\mathcal{T}^\mathrm{test}_1$   & $\mathcal{T}^\mathrm{train}_2$    & $\mathcal{T}^\mathrm{test}_2$   & $\mathcal{T}_1$   &  $\mathcal{T}^\mathrm{train}_3$    & $\mathcal{T}^\mathrm{test}_3$   & $\mathcal{T}_{1,2}$    & $\mathcal{T}^\mathrm{train}_1$    & $\mathcal{T}^\mathrm{test}_1$   \\
    Horizon & $200$ & $200$ & $400$ & $400$ & $400$ & $500$ & $500$ & $500$ & $5000$ & $5000$ \\
    \midrule
    GNN~\cite{battaglia2018gnn} & $0.74_{\pm.03}$ & $0.08_{\pm.06}$ & $0.55_{\pm.11}$ & $0.12_{\pm.10}$ & $0.00_{\pm.00}$ & $0.56_{\pm.15}$ & $0.07_{\pm.06}$ & $0.03_{\pm.02}$ & $0.39_{\pm.12}$ & $0.00_{\pm.00}$ \\
    TF~\cite{vaswani2017tf} & $0.74_{\pm.03}$ & $0.05_{\pm.10}$ & $0.66_{\pm.05}$ & $0.11_{\pm.12}$ & $0.68_{\pm.02}$ & \textbf{0.72}$_{\pm.05}$ & $0.08_{\pm.06}$ & $0.35_{\pm.16}$ & $0.15_{\pm.14}$ & $0.00_{\pm.00}$ \\
    VLM-OS~\cite{openai2026gpt54thinking} & $0.74_{\pm.01}$ & $0.04_{\pm.03}$ & $0.19_{\pm.15}$ & $0.02_{\pm.01}$ & $0.21_{\pm.10}$ & $0.00_{\pm.00}$ & $0.09_{\pm.08}$ & $0.23_{\pm.14}$ & $0.26_{\pm.02}$ & $0.00_{\pm.00}$ \\
    DSG-S~\cite{bagaria2025im-dsg} & $0.66_{\pm.05}$ & $0.04_{\pm.03}$ & $0.52_{\pm.14}$ & $0.01_{\pm.01}$ & $0.35_{\pm.19}$ & $0.68_{\pm.04}$ & $0.00_{\pm.00}$ & $0.17_{\pm.07}$ & $0.35_{\pm.03}$ & $0.00_{\pm.00}$ \\
    DSG-M~\cite{bagaria2025im-dsg}   & $0.38_{\pm.06}$ & $0.05_{\pm.03}$ & $0.01_{\pm.02}$ & $0.01_{\pm.01}$ & $0.07_{\pm.10}$ & $0.00_{\pm.00}$ & $0.05_{\pm.03}$ & $0.05_{\pm.02}$ & $0.18_{\pm.03}$ & $0.15_{\pm.00}$ \\
    RC~\cite{vats2024recoverychaining} & \textbf{0.75}$_{\pm.03}$ & $0.04_{\pm.04}$ & $0.61_{\pm.11}$ & $0.24_{\pm.18}$ & $0.39_{\pm.17}$ & $0.10_{\pm.03}$ & $0.16_{\pm.12}$ & $0.17_{\pm.14}$ & $0.00_{\pm.00}$ & $0.00_{\pm.00}$ \\
    RC+IVNTR~\cite{li2025IVNTR} & $0.74_{\pm.02}$ & $0.70_{\pm.03}$ & $0.64_{\pm.04}$ & $0.17_{\pm.13}$ & $0.72_{\pm.02}$ & $0.64_{\pm.04}$ & $0.37_{\pm.17}$ & $0.09_{\pm.10}$ & $0.47_{\pm.12}$ & $0.48_{\pm.12}$ \\
    \textbf{ReSYNC} & $0.73_{\pm.04}$ & $\textbf{0.76}_{\pm.05}$ & $\textbf{0.67}_{\pm.10}$ & $\textbf{0.84}_{\pm.06}$ & $\textbf{0.76}_{\pm.02}$ & $0.69_{\pm.05}$ & $\textbf{0.72}_{\pm.08}$ & $\textbf{0.80}_{\pm.06}$ & $\textbf{0.59}_{\pm.03}$ & $\textbf{0.57}_{\pm.03}$ \\
    \bottomrule[1.5pt]
    \end{tabular}%
  \caption{Success rate comparison on the four simulated domains.
  Our \model{} achieves the best results in unseen tests and competitive results in seen tasks.
  \model{} also progressively learns from multiple stages without forgetting.
  Results are averaged over five random seeds ($\pm$ stdev).
  }
  \label{tab:empirical}%
  \vspace{-1em}
\end{table*}%

\begin{table*}[!t]
    \centering
    \setlength{\tabcolsep}{1.4mm}
    \fontsize{7.5}{8}\selectfont
    \begin{tabular}{l|cccc|cccc|cccc|cccc}
    \toprule[1.5pt]
    Tasks & \multicolumn{8}{c|}{Cornered Insertion-S ($\mathcal{T}^\mathrm{train}_1|\mathcal{T}^\mathrm{test}_1$)}                     & \multicolumn{8}{c}{Cornered Insertion-L ($\mathcal{T}^\mathrm{train}_1|\mathcal{T}^\mathrm{test}_1$}) \\
    \midrule
    Seed  & S0    & S1    & S2    & AVG   & S0    & S1    & S2    & AVG   & S0    & S1    & S2    & AVG   & S0    & S1    & S2    & AVG \\
    Human & 0.9   & 0.7   & 0.8   & 0.80  & 0.7   & 0.9   & 0.8   & 0.80  & 0.8   & 0.8   & 0.6   & 0.73  & 0.8   & 0.7   & 0.8   & 0.77 \\
    \midrule
    RC    & 0.8   & 0.8   & 0.7   & \textbf{0.77}  & 0.0   & 0.0   & 0.0   & 0.00  & 0.7   & 0.7   & 0.5   & 0.63  & 0.0   & 0.0   & 0.0   & 0.00 \\
    \textbf{ReSYNC} & 0.7   & 0.8   & 0.7   & 0.73  & 0.6   & 0.8   & 0.9   & \textbf{0.77}  & 0.6   & 0.8   & 0.6   & \textbf{0.67}  & 0.5   & 0.6   & 0.8   & \textbf{0.63} \\
    \bottomrule[1.5pt]
    \end{tabular}%
  \caption{Quantitative results for real world domains. In $\mathcal{T}^\mathrm{train}_1$, both \model{} and RC~\cite{vats2024recoverychaining} perform comparably. In $\mathcal{T}^\mathrm{test}_1$, \model{} uses learned predicates to avoid failure, yielding better results.}
  \label{tab:real}%
  \vspace{-2em}
\end{table*}%

\myparagraph{Domains:}
We study four simulated domains and two real-world domains (\fref{fig:example}, \fref{fig:domains}) involving long-horizon physical interactions; details are deferred to \appref{app:domain_details}.
\begin{tightlist}
    \item \textit{Icy Transport (Sim):} A 2D cargo transport task~\cite{huang2026kinder} where icy or muddy doorway regions induce stochastic disturbances, requiring stable transport and generalization to unseen layouts.

    \item \textit{Blocked Stacking (Sim):} A 2D robot~\cite{huang2026kinder} stacks blocks while clearing static and movable obstructions.

    \item \textit{Cluttered Drawer (Sim):} A 3D mobile manipulator retrieves a hammer from cluttered scenes with drawers and obstructions, implemented in ManiSkill~\cite{shuklamaniskill}.

    \item \textit{Rearrange (Sim):} Adapted from ManiSkill-HAB~\cite{shuklamaniskill}, this domain requires rearranging objects despite several large obstructions at target locations, demanding tool-use skills.

    \item \textit{Cornered Insertion-S (Real):} A UR7e inserts Lego blocks into a tower. One block begins trapped in a box corner, requiring learned non-prehensile manipulation before insertion.

    \item \textit{Cornered Insertion-L (Real):} A longer version of ``Cornered Insertion-S", where the robot additionally needs to remove a lid from the box and place the tower into a closed drawer.
\end{tightlist}

\myparagraph{Experiment Setup:}
Each domain follows a staged curriculum for skill and concept acquisition.
After each stage, we evaluate on: (i) $\mathcal{T}_i^\mathrm{train}$ tasks sampled from the training distribution;
(ii) $\mathcal{T}_i^\mathrm{test}$ tasks generated by swapping object-centric states among same-type objects;
and (iii) $\mathcal{T}_{i-1}$ tasks from earlier stages.
We report mean success rates and standard deviations over 5 seeds, averaged across 50 tasks per scenario and all scenarios within each category; detailed results and explanations are in \appref{app:detailed_succ}.
For real-world domains, we follow OmniReset~\cite{yin2026omnireset} to train recovery policies in IsaacSim with domain randomization~\cite{bjelonic2025towards,akkaya2019solving}, then distill them into RGB policies~\cite{chi2023diffusionpolicy} using calibrated three-view observations.
Predicates are learned from simulated object poses.
The robot initially scans the scene with its wrist camera and SAM3~\cite{carion2025sam} to obtain object-centric states and generates a high-level plan with learned predicates, and then sequentially executes learned and given skills.
For both $\mathcal{T}_i^\mathrm{train}$ and $\mathcal{T}_i^\mathrm{test}$ scenarios, we evaluate 10 randomly initialized tasks over 3 seeds.

\subsection{Empirical Results}

\myparagraph{Simulated Domains:} We report average success rates across all four domains in \tref{tab:empirical}.
On $\mathcal{T}_i^\mathrm{train}$ tasks, \model{} successfully recovers and performs competitively with or better than the baselines (\textbf{Q1}).
\model{} shows three strengths: (1) by leveraging learned relational abstractions, \model{} significantly outperforms baselines by $20-70\%$ on $\mathcal{T}_i^\mathrm{test}$ scenarios (\textbf{Q1}).
(2) \model{} is able to learn from multiple training stages without a significant performance drop as task horizon and variance grow (\textbf{Q2}).
(3) \model{} does not forget the previous tasks when progressively learning from new experiences (\textbf{Q2}).
We present detailed analysis of baseline failures in \appref{app:baseline_details}.

\begin{wraptable}{t}{0.4\linewidth}
    \tablestyle{1.7pt}{0.8}
    \scriptsize
    \fontsize{7.5}{8}\selectfont
    \vspace{-0.6cm}
    \begin{center}
        \begin{tabular}{c|cc|ccc}
        \toprule[1.5pt]
        Metric & $J_0^{\times1e5}$ ($\downarrow$)    & $J_{-1}^{\times1e5}$    & $\mathcal{T}_2^\mathrm{train}$    & $\mathcal{T}_2^\mathrm{test}$   & $\mathcal{T}_1$ \\
        \midrule
        \xmark    & 2.47  & 2.04  & 0.28  & 0.47  & 0.76 \\
        \cmark   & 0.48  & 0.03  & 0.67  & 0.84  & 0.76 \\
        $\Delta$ & $\downarrow0.81$  & $\downarrow0.99$  & $\uparrow0.39$  & $\uparrow0.37$  & $0.00$ \\
        \bottomrule[1.5pt]
        \end{tabular}%
    \end{center}
    \vspace{-0.3cm}
    \caption{Ablation of concept fine-tuning in Blocked Stacking. Without fine-tuning, planning objectives inflate and training- and test-distribution success rates drop by 39 and 37 percentage points, respectively.}
    \label{tab:fine-tuning}
    \vspace{-0.1in}
\end{wraptable}

\myparagraph{Real Domains:}
As shown in \fref{fig:real}, \model{} learns a non-prehensile recovery skill from failed picking attempts and discovers new predicates that ground this skill.
These predicates allow \model{} to anticipate that ``Insert" will fail unless the block is first repositioned, enabling proactive recovery before insertion.
In contrast, reactive recovery~\cite{vats2024recoverychaining} waits until failure occurs, at which point recovery can no longer restore a solvable state.
Thus, although both methods recover from the training failure, only \model{} generalizes the learned skill as a proactive planning operator (\tref{tab:real}).
For reference, we also report a ``Human" baseline in which a human specifies the next skill after each previous skill.
Representative failure cases are shown in \appref{app:failure}.
These experiments demonstrate real-world abstract planning with a learned non-prehensile RL skill and discovered predicates (\textbf{Q3}).

\subsection{Ablations and Analysis}

\myparagraph{Relational Data Augmentation:}
In \tref{tab:empirical}, RC+IVNTR is trained with the same number of trajectories as $\mathcal{D}^{\mathcal{N}}$, but drawn solely from recovery experience.
In the first training stage across four domains, predicate learning without relational data augmentation results in a $10$--$36\%$ performance drop.
In contrast, relational data augmentation in \model{} provides a diverse trajectory distribution, yielding generalizable abstractions (\textbf{Q4}).

\begin{wrapfigure}{r}{0.5\columnwidth}
  \begin{center}
  \vspace{-0.3in}
    \includegraphics[width=0.5\columnwidth]{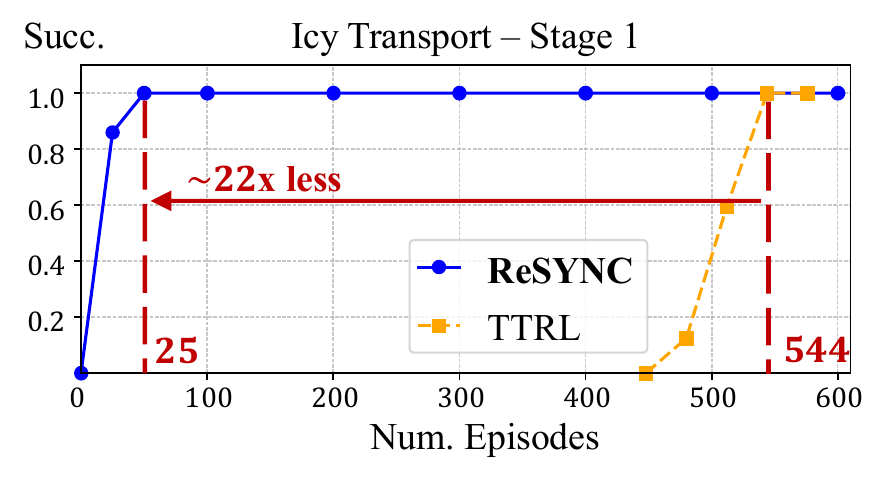}
    \vspace{-0.2in}
    \caption{Learning efficiency comparison. Thanks to relational abstractions and planning, \model{} reduces training episodes by $\sim 22\times$.}
    \label{fig:learning-efficiency}
    \vspace{-0.2in}
  \end{center}
\end{wrapfigure}

\myparagraph{Finetuning Across Scenarios:}
As demonstrated in \tref{tab:fine-tuning}, omitting fine-tuning leads to ``semantic drift,'' where previously learned predicates return erroneous truth values in the new states.
This results in higher planning objectives ($J$) and lower success rates (\textbf{Q4}).

\myparagraph{Sample Efficiency:} We compare \model{} with Test-Time Recovery Learning (TTRL), which learns new skills for new failures at test time.
As shown in \fref{fig:learning-efficiency}, \model{} solves $\mathcal{T}_i^\mathrm{train}$ tasks with $22\times$ fewer episodes by reusing learned concepts for symbolic reasoning (\textbf{Q5}).
We present more results about test-time planning efficiency comparison in \appref{app:plan_eff}.

\myparagraph{Additional Analysis:} We evaluate robustness to curriculum order and imperfect initial knowledge in \appref{app:revsed_order} and \appref{app:imperfect}.
\appref{app:additional_experiments} reports additional experiments on rejection-based data augmentation, contact-rich task, and visual identification of failure-causing objects.

\section{Limitations and Conclusion}

\myparagraph{Limitations:}
\model{} relies on a user-specified failure-eliciting curriculum, which could be automated with the recent agentic learning frameworks~\cite{liu2025lifelong,fu2026cap,zabounidis2026scalar}.
Our default relational data augmentation uses pairwise Gaussian state distributions; preliminary rejection-sampling results show that this choice can be relaxed, but the alternative is slow and retains noisy states (\appref{app:additional_experiments}).
More expressive generative models~\cite{luo2025concept,luo2026dictionary} could improve state coverage.
Object-centric state projection can omit relevant context, and our framework still assumes failure detection and identification of responsible objects; the visual identification experiment is preliminary.
\model{} inherits the limitations of IVNTR~\cite{li2025IVNTR}, including limited expressiveness and one-to-one skill-operator mappings.
See \appref{app:assumptions} for more discussions.

\myparagraph{Conclusion:}
We presented \model{}, a method that jointly learns skills and concepts for abstract planning from failure-recovery interaction.
Specifically, \model{} integrates recovery-skill acquisition with compositional relational augmentation and concept discovery, turning local reactive recovery into proactive task-level planning.
Across four physical domains, \model{} substantially improves compositional generalization and sample efficiency over various prior baselines.
We further demonstrate sim-to-real transfer of \model{}, where it generalizes the usage of the learned non-prehensile recovery skill via discovered predicates and abstract planning.

\acknowledgments{We acknowledge the support of the Air Force Research Laboratory (AFRL), DARPA, under agreement number FA8750-23-2-1015.
The authors also acknowledge the support of Fujitsu, under agreement number OSP00020653.
This work was supported by a hardware grant from Nvidia.
We also acknowledge a Princeton SEAS Innovation grant and support from  the NVIDIA
Academic Grant Program.
The authors would also like to express sincere gratitude to Patrick Yin (UW) for his timely support and help on our real robot domain and sim2real experiments.
We also thank Jiayuan Mao (UPenn) for her helpful discussions at the early stages of this work.

}


\bibliography{example_paper}  

\newpage
\appendix
\section{Notation Table}\label{app:notation}

In \tref{tab:notation}, we present the main symbols used in \sref{sec:bg}, \sref{sec:method}, and \sref{app:resync_details}.

\begin{table}[htbp]
\centering
\small
\caption{Important notation used in this work.}
\label{tab:notation}
\begin{tabularx}{\linewidth}{@{}lX@{}}
\toprule
\textbf{Symbol} & \textbf{Description} \\
\midrule
$\mathbf{x} \in \mathcal{X}$ & Continuous world state; state space \\
$\mathbf{x}_{\mathtt{o}}$ & Feature vector of object $\mathtt{o}$ \\
$\mathbf{x}_0$ & Initial state of a task \\
$\mathcal{O}$ & Set of objects in the environment \\
$\mathtt{o}$ & Individual object \\
$\mathbf{a} \in \mathcal{A}$ & Low-level action; action space \\
$f(\mathbf{x}' \mid \mathbf{x}, \mathbf{a})$ & Environment dynamics \\
$P(\mathbf{x}_0)$ & Initial state distribution \\
$g \subseteq \mathcal{X}$ & Goal specification \\
$T = \langle \mathcal{O}, \mathbf{x}_0, g \rangle$ & Task definition \\
$\mathcal{T}_i^{\mathtt{train}}$ & Training tasks at curriculum stage $i$ \\
$N_i$ & Number of training tasks in stage $i$ \\
$\psi \in \Psi$ & Lifted predicate (concept) \\
$\underline{\psi}$ & Ground predicate \\
$\theta_\psi$ & Predicate classifier \\
$\mathtt{C} \in \mathcal{C}$ & Lifted skill \\
$\underline{\mathtt{C}}$ & Ground skill \\
$\pi_{\mathtt{C}}$ & Skill policy \\
$\beta_{\mathtt{C}}$ & Skill termination function \\
$\mathtt{Op}^{\mathcal{C}}$ & Operator set induced from skills \\
$\mathrm{Plan}_f(\cdot)$ & Integrated planning and execution procedure \\
$i$ & Curriculum stage index \\
$T_j$ & $j$-th training task in a stage \\
$\mathbf{x}^{\mathcal{F}}_j$ & State immediately preceding failure in $T_j$ \\
$\mathbf{x}'$ & Recovery target in skill learning; next state in transition tuples \\
$\underline{\mathtt{C}}^{\mathcal{F}}$ & Ground skill active at failure \\
$\mathcal{C}_i$ & Skill set after stage $i$ \\
$\Psi_i$ & Predicate set after stage $i$ \\
$\mathtt{o}^{\mathcal{F}}$ & Failure-causing object \\
$\mathtt{o}^{\mathcal{R}}$ & Recovery-relevant object \\
$\mathcal{D}^{\mathcal{F}}$ & Mined failure-state dataset \\
$\mathbf{x}'_0$ & Resampled initial state for mining/relational data augmentation \\
$\mathtt{C}^{\mathcal{N}}$ & Newly learned recovery skill \\
$\mathcal{C}'$ & Skill set after adding $\mathtt{C}^{\mathcal{N}}$ \\
$\mathrm{proj}(\mathbf{x})$ & Projection onto relevant objects \\
$\mathcal{D}^{\mathcal{N}}$ & Dataset for concept learning \\
$\mathcal{D}^{\mathcal{N}}_i$ & Concept-learning dataset at stage $i$ \\
$\tilde{\Psi}^{\mathcal{N}}$ & Candidate predicates discovered from augmented data \\
$\Psi^{\mathcal{N}}$ & Selected newly discovered predicates \\
$\Psi'$ & Expanded predicate set after learning \\
$\mathtt{Op}^{\mathcal{C}'}$ & Operator set after skill expansion \\
$\Delta^\psi$ & Lifted effect vector for predicate $\psi$ \\
$J$ & Planning-efficiency surrogate objective \\
\bottomrule
\end{tabularx}
\end{table}
\clearpage

\section{Detailed Success Rates in Scenarios}\label{app:detailed_succ}
We report detailed success rates for each testing scenario in \tref{tab:bstacking1}, \tref{tab:bstacking2}, \tref{tab:transport}, and \tref{tab:3d}.
Within each domain, scenarios are assigned unique IDs for ease of reference.
As shown in the tables, \model{} generalizes effectively to unseen tasks (highlighted in \colorbox[rgb]{.816, .808, .808}{dark gray}).
Moreover, due to its modular design, \model{} retains performance on tasks from earlier curriculum stages (highlighted in \colorbox[rgb]{.949, .949, .949}{light gray}), indicating minimal forgetting.

\begin{itemize}
    \item \textbf{Blocked Stacking.}
    Scenario IDs describe which obstruction configurations are present while the robot stacks blocks.
    ID 0 contains the first-stage obstruction: a small fixed rectangle above the block to be grasped.
    ID 1 moves the first-stage obstruction to near target bottom block.
    ID 2 combines the first-stage obstruction with the second-stage movable large obstruction in a training arrangement, where both obstructions appear near the block to be grasped.
    IDs 3--5 are held-out combinations of the first- and second-stage obstructions, moving them to the target bottom block.
    IDs 6--7 isolate second-stage tasks with the movable large obstruction only.
    At the third stage, ID 8 is the training task involving all three obstructions, where the new thin vertical wall is placed near the bottom target block and the previous two obstructions are near the block to be grasped.
    ID 9 involves all the three obstructions but with the thin vertical wall moved to near the block to be grasped.
    IDs 10--13 are held-out combinations involving the newly introduced thin vertical wall with one of the earlier obstructions, applied to either the block to be grasped or the target bottom block.
    IDs 14--15 isolate the third-stage wall obstruction.

    \item \textbf{Icy Transport.}
    Scenario IDs describe which doorway perturbations the cargo robot must handle while transporting objects between rooms.
    Note that in this domain, the provided tasks in the same stage could involve different relevant objects.
    IDs 0--1 are first-stage tasks with icy doorway regions in front of one of the two transport objects.
    ID 2 is the held-out icy-layout scenario, where the icy region is in front of the transport target region.
    IDs 3--6 are the seen tasks combining first-stage icy-region with second-stage muddy-region, which are in front of one of the transport objects.
    IDs 7--11 are held-out mixed scenes where one of the icy/muddy regions appear in front of the transport target region, while the other appears in front of one of the transport objects.

    \item \textbf{Cluttered Drawer.}
    Scenario IDs describe drawer-and-clutter configurations for retrieving the hammer.
    ID 0 is the first-stage scene where the drawer is not fully open and must be pulled before the hammer can be reached.
    ID 1 is the held-out version, where the robot needs to place the hammer into the closed drawer.
    ID 2 is the second-stage training scene that combines the drawer condition with a movable block near the hammer.
    IDs 3--5 are held-out second-stage scenes where the block and drawer could be around the target placement region.

    \item \textbf{Rearrange.}
    Scenario IDs describe mobile-manipulation rearrangement scenes with large unpickable obstructions near target locations.
    ID 0 is the first-stage training distribution, where the obstructions are near the goal of the yellow-white sugar box.
    ID 1 is the held-out rearrangement scene, where the same obstructions are near the goal of the red bowl.
\end{itemize}

\begin{table*}[!t]
    \centering
    \setlength{\tabcolsep}{1.2mm}
    \fontsize{8}{11}\selectfont
    \begin{tabular}{c|cc|cccccccc}
    \toprule[1.5pt]
    Stages & \multicolumn{2}{c|}{$i=1$} & \multicolumn{8}{c}{$i=2$} \\
    \midrule
    Scenario ID & 0     & \cellcolor[rgb]{ .816,  .808,  .808}1 & 2     & \cellcolor[rgb]{ .949,  .949,  .949}0 & \cellcolor[rgb]{ .949,  .949,  .949}1 & \cellcolor[rgb]{ .816,  .808,  .808}3 & \cellcolor[rgb]{ .816,  .808,  .808}4 & \cellcolor[rgb]{ .816,  .808,  .808}5 & \cellcolor[rgb]{ .816,  .808,  .808}6 & \cellcolor[rgb]{ .816,  .808,  .808}7 \\
    \model{}  & 0.73  & \cellcolor[rgb]{ .816,  .808,  .808}\textbf{0.76} & \textbf{0.67} & \cellcolor[rgb]{ .949,  .949,  .949}\textbf{0.74} & \cellcolor[rgb]{ .949,  .949,  .949}\textbf{0.77} & \cellcolor[rgb]{ .816,  .808,  .808}\textbf{0.72} & \cellcolor[rgb]{ .816,  .808,  .808}\textbf{0.75} & \cellcolor[rgb]{ .816,  .808,  .808}\textbf{0.73} & \cellcolor[rgb]{ .816,  .808,  .808}\textbf{1.00} & \cellcolor[rgb]{ .816,  .808,  .808}\textbf{1.00} \\
    RC    & 0.75  & \cellcolor[rgb]{ .816,  .808,  .808}0.04 & 0.61  & \cellcolor[rgb]{ .949,  .949,  .949}0.75 & \cellcolor[rgb]{ .949,  .949,  .949}0.04 & \cellcolor[rgb]{ .816,  .808,  .808}0.03 & \cellcolor[rgb]{ .816,  .808,  .808}0.00 & \cellcolor[rgb]{ .816,  .808,  .808}0.00 & \cellcolor[rgb]{ .816,  .808,  .808}1.00 & \cellcolor[rgb]{ .816,  .808,  .808}0.16 \\
    RC+IVNTR & 0.74 & \cellcolor[rgb]{ .816,  .808,  .808}0.70 & 0.64 & \cellcolor[rgb]{ .949,  .949,  .949}0.74 & \cellcolor[rgb]{ .949,  .949,  .949}0.70 & \cellcolor[rgb]{ .816,  .808,  .808}0.00 & \cellcolor[rgb]{ .816,  .808,  .808}0.00 & \cellcolor[rgb]{ .816,  .808,  .808}0.61 & \cellcolor[rgb]{ .816,  .808,  .808}0.00 & \cellcolor[rgb]{ .816,  .808,  .808}0.23 \\
    VLM-OS & 0.74 & \cellcolor[rgb]{ .816,  .808,  .808}0.04 & 0.02 & \cellcolor[rgb]{ .949,  .949,  .949}0.38 & \cellcolor[rgb]{ .949,  .949,  .949}0.04 & \cellcolor[rgb]{ .816,  .808,  .808}0.01 & \cellcolor[rgb]{ .816,  .808,  .808}0.00 & \cellcolor[rgb]{ .816,  .808,  .808}0.00 & \cellcolor[rgb]{ .816,  .808,  .808}0.82 & \cellcolor[rgb]{ .816,  .808,  .808}0.10 \\
    DSG-M   & 0.38  & \cellcolor[rgb]{ .816,  .808,  .808}0.05 & 0.01  & \cellcolor[rgb]{ .949,  .949,  .949}0.32 & \cellcolor[rgb]{ .949,  .949,  .949}0.04 & \cellcolor[rgb]{ .816,  .808,  .808}0.00 & \cellcolor[rgb]{ .816,  .808,  .808}0.00 & \cellcolor[rgb]{ .816,  .808,  .808}0.00 & \cellcolor[rgb]{ .816,  .808,  .808}0.03 & \cellcolor[rgb]{ .816,  .808,  .808}0.06 \\
    GNN & 0.74  & \cellcolor[rgb]{ .816,  .808,  .808}0.08 & 0.55  & \cellcolor[rgb]{ .949,  .949,  .949}0.00 & \cellcolor[rgb]{ .949,  .949,  .949}0.00 & \cellcolor[rgb]{ .816,  .808,  .808}0.07 & \cellcolor[rgb]{ .816,  .808,  .808}0.17 & \cellcolor[rgb]{ .816,  .808,  .808}0.35 & \cellcolor[rgb]{ .816,  .808,  .808}0.00 & \cellcolor[rgb]{ .816,  .808,  .808}0.00 \\
    TF & 0.74  & \cellcolor[rgb]{ .816,  .808,  .808}0.05 & 0.66  & \cellcolor[rgb]{ .949,  .949,  .949}0.72 & \cellcolor[rgb]{ .949,  .949,  .949}0.77 & \cellcolor[rgb]{ .816,  .808,  .808}0 & \cellcolor[rgb]{ .816,  .808,  .808}0.34 & \cellcolor[rgb]{ .816,  .808,  .808}0.01 & \cellcolor[rgb]{ .816,  .808,  .808}0.2 & \cellcolor[rgb]{ .816,  .808,  .808}0.00 \\
    \bottomrule[1.5pt]
    \end{tabular}%
  \caption{Detailed success rate in each scenario after the first and the second learning stage in the Blocked Stacking domain.}
  \label{tab:bstacking1}%
\end{table*}%

\begin{table*}[!t]
    \centering
    \setlength{\tabcolsep}{0.8mm}
    \fontsize{8}{11}\selectfont
    \begin{tabular}{c|cccccccccccccccc}
    \toprule[1.5pt]
    Stages & \multicolumn{16}{c}{$i=3$} \\
    \midrule
    Scenario ID & 8     & \cellcolor[rgb]{ .949,  .949,  .949}0 & \cellcolor[rgb]{ .949,  .949,  .949}1 & \cellcolor[rgb]{ .949,  .949,  .949}2 & \cellcolor[rgb]{ .949,  .949,  .949}3 & \cellcolor[rgb]{ .949,  .949,  .949}4 & \cellcolor[rgb]{ .949,  .949,  .949}5 & \cellcolor[rgb]{ .949,  .949,  .949}6 & \cellcolor[rgb]{ .949,  .949,  .949}7 & \cellcolor[rgb]{ .816,  .808,  .808}9 & \cellcolor[rgb]{ .816,  .808,  .808}10 & \cellcolor[rgb]{ .816,  .808,  .808}11 & \cellcolor[rgb]{ .816,  .808,  .808}12 & \cellcolor[rgb]{ .816,  .808,  .808}13 & \cellcolor[rgb]{ .816,  .808,  .808}14 & \cellcolor[rgb]{ .816,  .808,  .808}15 \\
    \model{}  & 0.69  & \cellcolor[rgb]{ .949,  .949,  .949}\textbf{0.67} & \cellcolor[rgb]{ .949,  .949,  .949}0.74 & \cellcolor[rgb]{ .949,  .949,  .949}\textbf{0.77} & \cellcolor[rgb]{ .949,  .949,  .949}\textbf{0.72} & \cellcolor[rgb]{ .949,  .949,  .949}\textbf{0.76} & \cellcolor[rgb]{ .949,  .949,  .949}\textbf{0.72} & \cellcolor[rgb]{ .949,  .949,  .949}\textbf{1.00} & \cellcolor[rgb]{ .949,  .949,  .949}\textbf{1.00} & \cellcolor[rgb]{ .816,  .808,  .808}\textbf{0.58} & \cellcolor[rgb]{ .816,  .808,  .808}\textbf{0.67} & \cellcolor[rgb]{ .816,  .808,  .808}\textbf{0.63} & \cellcolor[rgb]{ .816,  .808,  .808}\textbf{0.68} & \cellcolor[rgb]{ .816,  .808,  .808}\textbf{0.70} & \cellcolor[rgb]{ .816,  .808,  .808}\textbf{0.90} & \cellcolor[rgb]{ .816,  .808,  .808}0.91 \\
    RC    & 0.10  & \cellcolor[rgb]{ .949,  .949,  .949}0.61 & \cellcolor[rgb]{ .949,  .949,  .949}0.75 & \cellcolor[rgb]{ .949,  .949,  .949}0.03 & \cellcolor[rgb]{ .949,  .949,  .949}0.03 & \cellcolor[rgb]{ .949,  .949,  .949}0.00 & \cellcolor[rgb]{ .949,  .949,  .949}0.00 & \cellcolor[rgb]{ .949,  .949,  .949}1.00 & \cellcolor[rgb]{ .949,  .949,  .949}0.16 & \cellcolor[rgb]{ .816,  .808,  .808}0.00 & \cellcolor[rgb]{ .816,  .808,  .808}0.00 & \cellcolor[rgb]{ .816,  .808,  .808}0.11 & \cellcolor[rgb]{ .816,  .808,  .808}0.06 & \cellcolor[rgb]{ .816,  .808,  .808}0.52 & \cellcolor[rgb]{ .816,  .808,  .808}0.24 & \cellcolor[rgb]{ .816,  .808,  .808}0.92 \\
    RC+IVNTR & 0.64 & \cellcolor[rgb]{ .949,  .949,  .949}0.10 & \cellcolor[rgb]{ .949,  .949,  .949}0.15 & \cellcolor[rgb]{ .949,  .949,  .949}0.31 & \cellcolor[rgb]{ .949,  .949,  .949}0.00 & \cellcolor[rgb]{ .949,  .949,  .949}0.00 & \cellcolor[rgb]{ .949,  .949,  .949}0.12 & \cellcolor[rgb]{ .949,  .949,  .949}0.00 & \cellcolor[rgb]{ .949,  .949,  .949}0.05 & \cellcolor[rgb]{ .816,  .808,  .808}0.26 & \cellcolor[rgb]{ .816,  .808,  .808}0.65 & \cellcolor[rgb]{ .816,  .808,  .808}0.37 & \cellcolor[rgb]{ .816,  .808,  .808}0.14 & \cellcolor[rgb]{ .816,  .808,  .808}0.10 & \cellcolor[rgb]{ .816,  .808,  .808}0.53 & \cellcolor[rgb]{ .816,  .808,  .808}0.54 \\
    VLM-OS & 0.00 & \cellcolor[rgb]{ .949,  .949,  .949}0.00 & \cellcolor[rgb]{ .949,  .949,  .949}0.29 & \cellcolor[rgb]{ .949,  .949,  .949}0.72 & \cellcolor[rgb]{ .949,  .949,  .949}0.00 & \cellcolor[rgb]{ .949,  .949,  .949}0.02 & \cellcolor[rgb]{ .949,  .949,  .949}0.00 & \cellcolor[rgb]{ .949,  .949,  .949}0.11 & \cellcolor[rgb]{ .949,  .949,  .949}0.67 & \cellcolor[rgb]{ .816,  .808,  .808}0.00 & \cellcolor[rgb]{ .816,  .808,  .808}0.09 & \cellcolor[rgb]{ .816,  .808,  .808}0.06 & \cellcolor[rgb]{ .816,  .808,  .808}0.37 & \cellcolor[rgb]{ .816,  .808,  .808}0.10 & \cellcolor[rgb]{ .816,  .808,  .808}0.00 & \cellcolor[rgb]{ .816,  .808,  .808}0.00 \\
    DSG-M   & 0.00  & \cellcolor[rgb]{ .949,  .949,  .949}0.00 & \cellcolor[rgb]{ .949,  .949,  .949}0.32 & \cellcolor[rgb]{ .949,  .949,  .949}0.04 & \cellcolor[rgb]{ .949,  .949,  .949}0.01 & \cellcolor[rgb]{ .949,  .949,  .949}0.00 & \cellcolor[rgb]{ .949,  .949,  .949}0.00 & \cellcolor[rgb]{ .949,  .949,  .949}0.03 & \cellcolor[rgb]{ .949,  .949,  .949}0.05 & \cellcolor[rgb]{ .816,  .808,  .808}0.00 & \cellcolor[rgb]{ .816,  .808,  .808}0.00 & \cellcolor[rgb]{ .816,  .808,  .808}0.01 & \cellcolor[rgb]{ .816,  .808,  .808}0.00 & \cellcolor[rgb]{ .816,  .808,  .808}0.00 & \cellcolor[rgb]{ .816,  .808,  .808}0.09 & \cellcolor[rgb]{ .816,  .808,  .808}0.24 \\
    GNN & 0.56  & \cellcolor[rgb]{ .949,  .949,  .949}0.00 & \cellcolor[rgb]{ .949,  .949,  .949}0.00 & \cellcolor[rgb]{ .949,  .949,  .949}0.00 & \cellcolor[rgb]{ .949,  .949,  .949}0.00 & \cellcolor[rgb]{ .949,  .949,  .949}0.07 & \cellcolor[rgb]{ .949,  .949,  .949}0.05 & \cellcolor[rgb]{ .949,  .949,  .949}0.00 & \cellcolor[rgb]{ .949,  .949,  .949}0.00 & \cellcolor[rgb]{ .816,  .808,  .808}0.08 & \cellcolor[rgb]{ .816,  .808,  .808}0.00 & \cellcolor[rgb]{ .816,  .808,  .808}0.00 & \cellcolor[rgb]{ .816,  .808,  .808}0.02 & \cellcolor[rgb]{ .816,  .808,  .808}0.32 & \cellcolor[rgb]{ .816,  .808,  .808}0.00 & \cellcolor[rgb]{ .816,  .808,  .808}0.00 \\
    TF & 0.72  & \cellcolor[rgb]{ .949,  .949,  .949}0.55 & \cellcolor[rgb]{ .949,  .949,  .949}0.13 & \cellcolor[rgb]{ .949,  .949,  .949}0.14 & \cellcolor[rgb]{ .949,  .949,  .949}0.43 & \cellcolor[rgb]{ .949,  .949,  .949}0.50 & \cellcolor[rgb]{ .949,  .949,  .949}0.59 & \cellcolor[rgb]{ .949,  .949,  .949}0.20 & \cellcolor[rgb]{ .949,  .949,  .949}0.28 & \cellcolor[rgb]{ .816,  .808,  .808}0.14 & \cellcolor[rgb]{ .816,  .808,  .808}0.12 & \cellcolor[rgb]{ .816,  .808,  .808}0.02 & \cellcolor[rgb]{ .816,  .808,  .808}0.16 & \cellcolor[rgb]{ .816,  .808,  .808}0.00 & \cellcolor[rgb]{ .816,  .808,  .808}0.00 & \cellcolor[rgb]{ .816,  .808,  .808}0.00 \\
    \bottomrule[1.5pt]
    \end{tabular}%
  \caption{Detailed success rate in each scenario after the third learning stage in the Blocked Stacking domain.}
  \label{tab:bstacking2}%
\end{table*}%

\begin{table*}[!t]
    \centering
    \setlength{\tabcolsep}{0.9mm}
    \fontsize{7.5}{10}\selectfont
    \begin{tabular}{c|ccc|cccccccccccc}
    \toprule[1.5pt]
    Stages & \multicolumn{3}{c|}{$i=1$} & \multicolumn{12}{c}{$i=2$} \\
    \midrule
    Scenario ID & 0     & 1     & \cellcolor[rgb]{ .816,  .808,  .808}2 & 3     & 4     & 5     & 6     & \cellcolor[rgb]{ .949,  .949,  .949}0 & \cellcolor[rgb]{ .949,  .949,  .949}1 & \cellcolor[rgb]{ .949,  .949,  .949}2 & \cellcolor[rgb]{ .816,  .808,  .808}7 & \cellcolor[rgb]{ .816,  .808,  .808}8 & \cellcolor[rgb]{ .816,  .808,  .808}9 & \cellcolor[rgb]{ .816,  .808,  .808}10 & \cellcolor[rgb]{ .816,  .808,  .808}11 \\
    \model{}  & 0.90  & 0.94  & \cellcolor[rgb]{ .816,  .808,  .808}\textbf{0.76} & \textbf{0.74} & \textbf{0.83} & \textbf{0.89} & \textbf{0.94} & \cellcolor[rgb]{ .949,  .949,  .949}\textbf{0.91} & \cellcolor[rgb]{ .949,  .949,  .949}\textbf{0.94} & \cellcolor[rgb]{ .949,  .949,  .949}\textbf{0.76} & \cellcolor[rgb]{ .816,  .808,  .808}\textbf{0.89} & \cellcolor[rgb]{ .816,  .808,  .808}\textbf{0.56} & \cellcolor[rgb]{ .816,  .808,  .808}\textbf{0.84} & \cellcolor[rgb]{ .816,  .808,  .808}\textbf{0.40} & \cellcolor[rgb]{ .816,  .808,  .808}\textbf{0.74} \\
    RC    & 0.60  & 0.82  & \cellcolor[rgb]{ .816,  .808,  .808}0.00 & 0.44  & 0.13  & 0.42  & 0.35  & \cellcolor[rgb]{ .949,  .949,  .949}0.74 & \cellcolor[rgb]{ .949,  .949,  .949}0.77 & \cellcolor[rgb]{ .949,  .949,  .949}0.00 & \cellcolor[rgb]{ .816,  .808,  .808}0.00 & \cellcolor[rgb]{ .816,  .808,  .808}0.00 & \cellcolor[rgb]{ .816,  .808,  .808}0.00 & \cellcolor[rgb]{ .816,  .808,  .808}0.00 & \cellcolor[rgb]{ .816,  .808,  .808}0.00 \\
    RC+IVNTR & 0.60 & 0.41 & \cellcolor[rgb]{ .816,  .808,  .808}0.44 & 0.18 & 0.14 & 0.26 & 0.14 & \cellcolor[rgb]{ .949,  .949,  .949}0.79 & \cellcolor[rgb]{ .949,  .949,  .949}0.57 & \cellcolor[rgb]{ .949,  .949,  .949}0.62 & \cellcolor[rgb]{ .816,  .808,  .808}0.11 & \cellcolor[rgb]{ .816,  .808,  .808}0.11 & \cellcolor[rgb]{ .816,  .808,  .808}0.02 & \cellcolor[rgb]{ .816,  .808,  .808}0.11 & \cellcolor[rgb]{ .816,  .808,  .808}0.04 \\
    VLM-OS & 0.20 & 0.57 & \cellcolor[rgb]{ .816,  .808,  .808}0.00 & 0.18 & 0.25 & 0.04 & 0.74 & \cellcolor[rgb]{ .949,  .949,  .949}0.32 & \cellcolor[rgb]{ .949,  .949,  .949}0.40 & \cellcolor[rgb]{ .949,  .949,  .949}0.00 & \cellcolor[rgb]{ .816,  .808,  .808}0.00 & \cellcolor[rgb]{ .816,  .808,  .808}0.00 & \cellcolor[rgb]{ .816,  .808,  .808}0.00 & \cellcolor[rgb]{ .816,  .808,  .808}0.00 & \cellcolor[rgb]{ .816,  .808,  .808}0.00 \\
    DSG-M   & 0.58  & 0.64  & \cellcolor[rgb]{ .816,  .808,  .808}0.33 & 0.00  & 0.00  & 0.37  & 0.22  & \cellcolor[rgb]{ .949,  .949,  .949}0.30 & \cellcolor[rgb]{ .949,  .949,  .949}0.23 & \cellcolor[rgb]{ .949,  .949,  .949}0.09 & \cellcolor[rgb]{ .816,  .808,  .808}0.07 & \cellcolor[rgb]{ .816,  .808,  .808}0.00 & \cellcolor[rgb]{ .816,  .808,  .808}0.00 & \cellcolor[rgb]{ .816,  .808,  .808}0.00 & \cellcolor[rgb]{ .816,  .808,  .808}0.00 \\
    GNN & 1.00  & 0.96  & \cellcolor[rgb]{ .816,  .808,  .808}0.00 & 0.00  & 0.30  & 0.26  & 0.15  & \cellcolor[rgb]{ .949,  .949,  .949}0.30 & \cellcolor[rgb]{ .949,  .949,  .949}0.34 & \cellcolor[rgb]{ .949,  .949,  .949}0.50 & \cellcolor[rgb]{ .816,  .808,  .808}0.00 & \cellcolor[rgb]{ .816,  .808,  .808}0.09 & \cellcolor[rgb]{ .816,  .808,  .808}0.00 & \cellcolor[rgb]{ .816,  .808,  .808}0.27 & \cellcolor[rgb]{ .816,  .808,  .808}0.56 \\
    TF & 1.00  & 0.96  & \cellcolor[rgb]{ .816,  .808,  .808}0.00 & 0.10  & 0.16  & 0.32  & 0.45  & \cellcolor[rgb]{ .949,  .949,  .949}0.39 & \cellcolor[rgb]{ .949,  .949,  .949}0.41 & \cellcolor[rgb]{ .949,  .949,  .949}0.77 & \cellcolor[rgb]{ .816,  .808,  .808}0.03 & \cellcolor[rgb]{ .816,  .808,  .808}0.00 & \cellcolor[rgb]{ .816,  .808,  .808}0.00 & \cellcolor[rgb]{ .816,  .808,  .808}0.29 & \cellcolor[rgb]{ .816,  .808,  .808}0.52 \\
    \bottomrule[1.5pt]
    \end{tabular}%
  \caption{Detailed success rate in each scenario after the first and the second learning stage in the Icy Transport domain.}
  \label{tab:transport}%
\end{table*}%

\begin{table*}[!t]
    \centering
    \setlength{\tabcolsep}{1.8mm}
    \fontsize{8.5}{12}\selectfont
    \begin{tabular}{ccc|cccccc|cc}
    \toprule[1.5pt]
    Scenario ID & \multicolumn{2}{c|}{Cluttered Drawer $i=1$} & \multicolumn{6}{c|}{Cluttered Drawer $i=2$}                     & \multicolumn{2}{c}{Rearrange $i=1$} \\
    \midrule
    Method & 0     & \cellcolor[rgb]{ .816,  .808,  .808}1 & 2     & \cellcolor[rgb]{ .949,  .949,  .949}0 & \cellcolor[rgb]{ .949,  .949,  .949}1 & \cellcolor[rgb]{ .816,  .808,  .808}3 & \cellcolor[rgb]{ .816,  .808,  .808}4 & \cellcolor[rgb]{ .816,  .808,  .808}5 & 0     & \cellcolor[rgb]{ .816,  .808,  .808}1 \\
    \model{}  & \textbf{0.95} & \cellcolor[rgb]{ .816,  .808,  .808}\textbf{0.96} & 0.64  & \cellcolor[rgb]{ .949,  .949,  .949}\textbf{0.74} & \cellcolor[rgb]{ .949,  .949,  .949}\textbf{0.78} & \cellcolor[rgb]{ .816,  .808,  .808}\textbf{0.71} & \cellcolor[rgb]{ .816,  .808,  .808}\textbf{0.72} & \cellcolor[rgb]{ .816,  .808,  .808}\textbf{0.74} & \textbf{0.59} & \cellcolor[rgb]{ .816,  .808,  .808}\textbf{0.57} \\
    RC    & 0.68  & \cellcolor[rgb]{ .816,  .808,  .808}0.01 & 0.62  & \cellcolor[rgb]{ .949,  .949,  .949}0.66 & \cellcolor[rgb]{ .949,  .949,  .949}0.03 & \cellcolor[rgb]{ .816,  .808,  .808}0.40 & \cellcolor[rgb]{ .816,  .808,  .808}0.03 & \cellcolor[rgb]{ .816,  .808,  .808}0.00 & 0.00  & \cellcolor[rgb]{ .816,  .808,  .808}0.00 \\
    RC+IVNTR & 0.69 & \cellcolor[rgb]{ .816,  .808,  .808}0.59 & 0.64 & \cellcolor[rgb]{ .949,  .949,  .949}0.57 & \cellcolor[rgb]{ .949,  .949,  .949}0.59 & \cellcolor[rgb]{ .816,  .808,  .808}0.00 & \cellcolor[rgb]{ .816,  .808,  .808}0.00 & \cellcolor[rgb]{ .816,  .808,  .808}0.81 & 0.47 & \cellcolor[rgb]{ .816,  .808,  .808}0.48 \\
    VLM-OS & 0.68 & \cellcolor[rgb]{ .816,  .808,  .808}0.29 & 0.18 & \cellcolor[rgb]{ .949,  .949,  .949}0.68 & \cellcolor[rgb]{ .949,  .949,  .949}0.31 & \cellcolor[rgb]{ .816,  .808,  .808}0.36 & \cellcolor[rgb]{ .816,  .808,  .808}0.08 & \cellcolor[rgb]{ .816,  .808,  .808}0.32 & 0.00 & \cellcolor[rgb]{ .816,  .808,  .808}0.26 \\
    DSG-M   & 0.13  & \cellcolor[rgb]{ .816,  .808,  .808}0.03 & 0.00  & \cellcolor[rgb]{ .949,  .949,  .949}0.16 & \cellcolor[rgb]{ .949,  .949,  .949}0.03 & \cellcolor[rgb]{ .816,  .808,  .808}0.00 & \cellcolor[rgb]{ .816,  .808,  .808}0.00 & \cellcolor[rgb]{ .816,  .808,  .808}0.00 & 0.18  & \cellcolor[rgb]{ .816,  .808,  .808}0.15 \\
    GNN   & 0.86  & \cellcolor[rgb]{ .816,  .808,  .808}0.00 & 0.69  & \cellcolor[rgb]{ .949,  .949,  .949}0.00 & \cellcolor[rgb]{ .949,  .949,  .949}0.00 & \cellcolor[rgb]{ .816,  .808,  .808}0.47 & \cellcolor[rgb]{ .816,  .808,  .808}0.62 & \cellcolor[rgb]{ .816,  .808,  .808}0.34 & 0.39  & \cellcolor[rgb]{ .816,  .808,  .808}0.00 \\
    TF    & 0.11  & \cellcolor[rgb]{ .816,  .808,  .808}0.00 & 0.03  & \cellcolor[rgb]{ .949,  .949,  .949}0.00 & \cellcolor[rgb]{ .949,  .949,  .949}0.00 & \cellcolor[rgb]{ .816,  .808,  .808}0.00 & \cellcolor[rgb]{ .816,  .808,  .808}0.00 & \cellcolor[rgb]{ .816,  .808,  .808}0.00 & 0.15  & \cellcolor[rgb]{ .816,  .808,  .808}0.00 \\
    \bottomrule[1.5pt]
    \end{tabular}%
  \caption{Detailed success rate in each scenario after each learning stage in the Cluttered Drawer and the Rearrange domain.}
  \label{tab:3d}%
\end{table*}%

\section{Baseline Details and Analysis}\label{app:baseline_details}

\myparagraph{Recovery Chaining (RC)~\cite{vats2024recoverychaining}.}
RC retains the initial symbolic planner but does not incorporate learned recovery skills into its planning operators.
At test time, plans are generated using the initial predicates and operators; when a failure detector is triggered, the corresponding learned recovery skill is executed.
For fairness, RC uses the same learned failure detectors and skill policies as \model{}.
Its main limitation is generalization: test-time failure states can differ substantially from those seen during recovery training, often making recovery ineffective.
In long-horizon domains such as Icy Transport and Rearrange, RC also suffers from compounding detector errors, since the failure detector is queried at every low-level step.
By contrast, \model{} converts the failed skill and its recovery into new planning operators, so the relevant detector is invoked only when that skill is executed, reducing compounding errors and enabling more reliable recovery execution.

\myparagraph{RC+IVNTR~\cite{vats2024recoverychaining,li2025IVNTR}.}
RC+IVNTR applies IVNTR after each RC-trained recovery skill to learn predicates and operators, but does not use relational data augmentation or concept finetuning.
Compared with RC, it can use learned recovery skills during abstract planning, improving test-time generalization and reducing detector compounding in the same way as \model{}.
However, because concept learning only uses trajectories from the original recovery context, the learned predicates are often insufficiently robust for out-of-distribution test states.
Moreover, without finetuning, predicates discovered in earlier stages may become misaligned with later skills, leading to planning failures in multi-stage domains.

\myparagraph{Deep Skill Graph (DSG-S, DSG-M)~\cite{bagaria2025im-dsg}.}
The official implementation builds a single graph over all training tasks in each stage (DSG-S).
This performs well on training-distribution tasks and often matches \model{}, but it cannot synthesize plan sequences absent from the training data, resulting in poor test-time generalization.
We therefore also implement a multi-task variant, DSG-M, which constructs planning graphs using PDDL operators and integrates each learned recovery skill into the planner.
Specifically, DSG-M introduces two predicates per skill: one for the skill's add effect and one for its precondition/delete effect.
We use the same neural architectures and augmented trajectory data as \model{}.
However, DSG-M restricts predicates to this fixed structure, limiting expressiveness and often producing inefficient plans with redundant skill sequences.
Additional quantitative results are provided in \appref{app:plan_eff}.

\myparagraph{VLM One-shot (VLM-OS)~\cite{openai2026gpt54thinking}.}
VLM-OS prompts GPT-5.4~\cite{openai2026gpt54thinking} to generate a plan for each task in a single forward pass, without additional training or finetuning.
The prompt includes the initial PDDL operators, while newly learned skill operators are given empty preconditions and effects; the VLM must infer their symbolic descriptions from in-context training examples.
An example prompt for the first stage of Blocked Stacking is shown in \fref{fig:incontex}.
VLM-OS can sometimes solve training-distribution tasks, but it struggles on test tasks because inferring precise symbolic descriptions from few examples is difficult.
This issue is amplified when planning with subsemantic RL skills whose effects are hard to express in natural language, consistent with prior observations~\cite{athalye2024predicate,kumar2024openworld}.

\begin{center}
    \begin{pikebox}{VLM-OS Prompt for Blocked Stacking}
    \begin{Verbatim}[fontsize=\tiny,breaklines=true]
You are an expert robot task planner. Given a scene image, a PDDL domain definition, and a PDDL problem, produce a ground operator plan that achieves the goal. The second image shows the current scene you must solve.

Carefully compare the two images to understand the objects, and their relationships, and the differences in initial states. Use the domain definition to understand the available actions, **the PDDL is incomplete** and requires you to infer the missing preconditions and effects based on the example and your understanding of the scene. Carefully inspect the image first before generating the output, **do not directly plan with the incomplete PDDL**

## PDDL Domain
```
(define (domain Domain_Scenario_1)
    (:requirements :typing)
    (:types
    block - dyn_rectangle
    obstruction_rec - kin_rectangle
    robot - kin_robot
    dynamic2d)

    (:predicates
    (Holding ?x0 - robot ?x1 - block)
    (On ?x0 - block ?x1 - block)
    (ReadyGrasp ?x0 - robot ?x1 - block)
    (ReadyPlace ?x0 - robot ?x1 - block ?x2 - block)
    )

    (:action Grasp
    :parameters (?robot - robot ?grasp_block - block)
    :precondition (and (ReadyGrasp ?robot ?grasp_block))
    :effect (and (Holding ?robot ?grasp_block)
        (not (ReadyGrasp ?robot ?grasp_block)))
    )

  (:action Place
    :parameters (?robot - robot ?grasp_block - block ?base_block - block)
    :precondition (and (Holding ?robot ?grasp_block)
        (ReadyPlace ?robot ?grasp_block ?base_block))
    :effect (and (On ?grasp_block ?base_block)
        (not (Holding ?robot ?grasp_block))
        (not (ReadyPlace ?robot ?grasp_block ?base_block)))
    )

  (:action Punch # New RL skill
    :parameters (?robot - robot ?block - block ?obstruction_rec - obstruction_rec)
    :precondition (and )
    :effect (and )
    )

  (:action ReachToGrasp
    :parameters (?robot - robot ?grasp_block - block)
    :precondition (and )
    :effect (and (ReadyGrasp ?robot ?grasp_block))
    )

  (:action ReachToGrasp_Punch # Previously failed skill
    :parameters (?robot - robot ?block - block ?obstruction_rec - obstruction_rec)
    :precondition (and )
    :effect (and )
    )

  (:action ReachToPlace
    :parameters (?robot - robot ?grasp_block - block ?base_block - block)
    :precondition (and (Holding ?robot ?grasp_block))
    :effect (and (ReadyPlace ?robot ?grasp_block ?base_block))
    )
)
```

## Solved Example
Here is a solved example from the same domain. The first image shows the example scene.

### Example Problem
```
(define (problem task)
  (:domain BlockedStacking2D-domain)
  (:objects base_block (red block) - block grasp_block (purple block) - block obstruction_rec (black small rectangle) - obstruction_rec robot (purple circle with arm and gripper) - robot)
  (:init )
  (:goal (and (On grasp_block base_block)))
)```

### Example Plan
```plan
(ReachToGrasp_Punch robot grasp_block obstruction_rec)
(Punch robot grasp_block obstruction_rec)
(ReachToGrasp robot grasp_block)
(Grasp robot grasp_block)
(ReachToPlace robot grasp_block base_block)
(Place robot grasp_block base_block)
```

## PDDL Problem
```
(define (problem task)
  (:domain Domain_Scenario_1)
  (:objects base_block (red block) - block grasp_block (purple block) - block obstruction_rec (black small rectangle) - obstruction_rec robot (purple circle with arm and gripper) - robot)
  (:init )
  (:goal (and (On grasp_block base_block)))
)```

## Output Format
Output ONLY the plan inside a ```plan``` code block, one action per line. Example:
```plan
(operator1 obj_a obj_b)
(operator2 obj_c obj_d)
```
    \end{Verbatim}
    \end{pikebox}
    \captionsetup{hypcap=false}\captionof{figure}{Example VLM-OS prompt for the Blocked Stacking domain. The prompt includes the incomplete PDDL domain, one solved example, the target problem, and the required plan-only output format.}
    \label{fig:incontex}
\end{center}

\myparagraph{GNN/TF Policy~\cite{battaglia2018gnn,vaswani2017tf}.}
GNN/TF policies are trained offline by supervised learning on segmented augmented trajectories, following previous work~\cite{li2025IVNTR,wang2025unipred,silver2023predicateinvent}.
Given a low-level state graph $G=(V,E,u)$, where nodes represent objects, edges encode pairwise relations, and $u$ denotes global scene features, the policy predicts both the high-level operator to execute and its object arguments.
Current-state and goal atoms from the initial concept set are concatenated to node and edge features to provide symbolic context.
Training minimizes a combined loss with cross-entropy for operator prediction and binary cross-entropy for object selection.
We evaluate two architectures.
The GNN~\cite{battaglia2018gnn} uses an Encode--Process--Decode design: MLPs embed node, edge, and global features, iterative message passing performs relational reasoning, and MLP decoders predict operator logits from the global embedding and object logits from node embeddings.
The Transformer~\cite{vaswani2017tf} flattens encoded node, edge, and global features into tokens, applies a multi-head self-attention encoder, and then unflattens the output to produce operator and object predictions with the same decoders.

\section{Domain Details}\label{app:domain_details}
To start, we introduce the hyper-parameters that are shared across all domains:
\begin{itemize}
    \item \textit{Neural Classifiers}: All of the neural classifiers (including all of the predicate classifiers and terminal functions) learned in this work are MLPs with $4$ hidden layers, with dimensions $128, 256, 256, 128$, respectively. We used GELU as activation function between layers and added 1D batch normalization. The classifiers are optimized with a learning rate of $0.001$ using AdamW as optimizer.
    \item \textit{Neural Policies}: All of the recovery skill policies (for all of the stages in all of the domains) take an Actor-Critic framework. Both the actor and the critic are modeled using MLPs with $4$ hidden layers, with dimensions $256, 256, 256, 256$, respectively. Tanh is applied as activation function. The policy is optimized using PPO with Adam as optimizer.
\end{itemize}
This section summarizes key different implementation details for each domain. Additional details are available in the accompanying code release.

\begin{enumerate}
\item \textbf{Icy Transport} \\
A 2D navigation domain inspired by Four Rooms~\cite{sutton1999between}, where a car-like robot transports objects between rooms through doorways. Doorways may contain icy or muddy regions that perturb the robot’s motion. Objects can be pushed via collision.
    \begin{itemize}
        \item \textit{State and actions:} States are represented using SE(2) poses of all objects. Actions are 3-dimensional vectors consisting of forward throttle, steering, and torque.
        \item \textit{Object types:} Robot ($\mathtt{r}$), transport object ($\mathtt{o}$), target ($\mathtt{t}$), icy region ($\mathtt{icy}$), muddy region ($\mathtt{muddy}$).
        \item \textit{Operators:} $\mathtt{GoToPickObject(?r, ?o)}$, $\mathtt{GoToPlaceObject(?r, ?o, ?t)}$.
        \item \textit{Predicates:} $\mathtt{HandEmpty(?r)}$, $\mathtt{Holding(?r, ?o)}$, $\mathtt{Delivered(?o, ?t)}$.
        \item \textit{Hyperparameters:} In both stages 1 and 2, recovery skills are trained with a horizon of $30$ and up to $500,000$ environment steps. For concept learning, we collect $600$ augmented trajectories in stage 1 and $1200$ in stage 2.
    \end{itemize}
    \begin{table}[htbp]
    \centering
    \footnotesize
    \setlength{\tabcolsep}{6pt}

    \label{tab:failure_Stages1}
    \begin{tabular}{c l l l}
    \toprule
    \textbf{Stages} & \textbf{Failure Objects} & \textbf{Failed Skill} & \textbf{New Skill} \\
    \midrule
    $i=1$ & Icy Region & $\mathtt{GoToPickObject}$ & $\mathtt{IcyDrive}$ \\
    $i=2$ & Muddy Region & $\mathtt{GoToPickObject}$ & $\mathtt{MuddyDrive}$ \\

    \bottomrule
    \end{tabular}
    \end{table}

\item \textbf{Blocked Stacking.} \\
A 2D manipulation domain in which a robot stacks blocks while manipulating around static and dynamic obstructions.
Obstruction shapes and property vary across curriculum stages.
\begin{itemize}
    \item \textit{State and actions:} States are represented using SE(2) poses of all objects. Actions are 5-dimensional vectors consisting of $\Delta x$, $\Delta y$, $\Delta \theta$, arm joint delta, and finger joint delta.
    \item \textit{Object types:} Robot ($\mathtt{r}$), blocks ($\mathtt{b}$), obstructions ($\mathtt{o}$).
    \item \textit{Operators:} $\mathtt{ReachToGrasp(?r, ?b)}$, $\mathtt{ReachToPlace(?r, ?b)}$, $\mathtt{Grasp(?r, ?b)}$, $\mathtt{Place(?r, ?b, ?b)}$.
    \item \textit{Predicates:} $\mathtt{ReadyGrasp(?r, ?b)}$, $\mathtt{ReadyPlace(?r, ?b, ?b)}$, $\mathtt{Holding(?r, ?b)}$, $\mathtt{On(?b, ?b)}$.
    \item \textit{Hyperparameters:} Recovery skills are trained with a horizon of $10$ and up to $500,000$ environment steps at all stages. For concept learning, we collect $1000$ augmented trajectories in stage 1, $2000$ in stage 2, and $6400$ in stage 3.
\end{itemize}

\begin{table}[htbp]
\centering
\footnotesize
\setlength{\tabcolsep}{6pt}

\label{tab:failure_Stages2}
\begin{tabular}{c l l l}
\toprule
\textbf{Stages} & \textbf{Failure Objects} & \textbf{Failed Skill} & \textbf{New Skill} \\
\midrule
$i=1$ & A small fixed rectangle above a block & $\mathtt{ReachToGrasp}$ & $\mathtt{Punch}$ \\
$i=2$ & A movable large obstruction next to a block & $\mathtt{ReachToGrasp}$ & $\mathtt{Wiggle}$ \\
$i=3$ & A fixed thin vertical wall leaning on a block & $\mathtt{ReachToGrasp}$ & $\mathtt{Fiddle}$  \\
\bottomrule
\end{tabular}
\end{table}

\item \textbf{Cluttered Drawer.} \\
A 3D mobile manipulation domain in ManiSkill, where a robot retrieves a hammer from cluttered scenes containing a closed drawer and movable blocks.
\begin{itemize}
    \item \textit{State and actions:} States are represented using SE(3) poses of all objects, augmented with articulation joint positions. Actions are 10-dimensional joint-position control commands for the mobile manipulator.
    \item \textit{Object types:} Robot ($\mathtt{r}$), hammer ($\mathtt{h}$), drawer ($\mathtt{d}$), block ($\mathtt{b}$).
    \item \textit{Operators:} $\mathtt{BodyReachToGrasp(?r, ?h)}$, $\mathtt{BodyReachToPlace(?r, ?h, ?h)}$, $\mathtt{HandReachToGrasp(?r, ?h)}$, $\mathtt{HandReachToPlace(?r, ?h, ?h)}$, $\mathtt{Grasp(?r, ?h)}$, $\mathtt{Place(?r, ?h, ?h)}$.
    \item \textit{Predicates:} $\mathtt{BodyReadyGrasp(?r, ?h)}$, $\mathtt{HandReadyGrasp(?r, ?h)}$, $\mathtt{BodyReadyPlace(?r, ?h, ?h)}$, $\mathtt{HandReadyPlace(?r, ?h, ?h)}$, $\mathtt{Holding(?r, ?h)}$, $\mathtt{On(?h, ?h)}$.
    \item \textit{Hyperparameters:} Recovery skills are trained with a horizon of $16$ and up to $5M$ environment steps at all stages. For concept learning, we collect $1000$ augmented trajectories in stage 1 and $1280$ in stage 2.
\end{itemize}
\begin{table}[htbp]
\centering
\footnotesize
\setlength{\tabcolsep}{6pt}

\label{tab:failure_Stages3}
\begin{tabular}{c l l l}
\toprule
\textbf{Stages} & \textbf{Failure Objects} & \textbf{Failed Skill} & \textbf{New Skill} \\
\midrule
$i=1$ & The drawer is not fully open & $\mathtt{HandReachToGrasp}$ & $\mathtt{Pull}$ \\
$i=2$ & A block next to a hammer & $\mathtt{HandReachToGrasp}$ & $\mathtt{Wiggle}$ \\
\bottomrule
\end{tabular}
\end{table}

\item \textbf{Rearrange.} \\
A 3D mobile manipulation domain based on ManiSkill-HAB, where a robot rearranges objects to target locations in the presence of large, unpickable obstructions.
\begin{itemize}
    \item \textit{State and actions:} States are represented using SE(3) poses of all objects. Actions are 10-dimensional joint-position control commands for the mobile manipulator.
    \item \textit{Object types:} Robot ($\mathtt{r}$), pick-and-place objects ($\mathtt{o}$), goal locations ($\mathtt{g}$), obstruction can ($\mathtt{can}$).
    \item \textit{Operators:} $\mathtt{GoToPickBowl(?r, ?o)}$, $\mathtt{PickBowl(?r, ?o)}$, $\mathtt{GoToPlaceBowl(?r, ?o, ?g)}$, $\mathtt{PlaceBowl(?r, ?o, ?g)}$, $\mathtt{GoToPickBox(?r, ?o)}$, $\mathtt{PickBox(?r, ?o)}$, $\mathtt{GoToPlaceBox(?r, ?o, ?g)}$, $\mathtt{PlaceBox(?r, ?o, ?g)}$.
    \item \textit{Predicates:} $\mathtt{IsBowl(?o)}$, $\mathtt{IsBox(?o)}$, $\mathtt{BodyReadyPick(?r, ?o)}$, $\mathtt{BodyReadyPlace(?r, ?g)}$, $\mathtt{HoldingBox(?r)}$, $\mathtt{HoldingBowl(?r)}$, $\mathtt{HandEmpty(?r)}$, $\mathtt{AtGoal(?o, ?g)}$, $\mathtt{BoxAtGoal(?r)}$, $\mathtt{NoBodyReadyPlace(?r)}$.
    \item \textit{Hyperparameters:} Recovery skills are trained with a horizon of $8$ and up to $5M$ environment steps at all stages. For concept learning, we collect 500 augmented trajectories in stage 1.
\end{itemize}

\item \textbf{Cornered Insertion-S.} \
A UR7e arm with a Robotiq 2F-140 gripper must insert one block onto another to form a tower. One block is initially placed near a box corner, causing the nominal insertion behavior to fail. Train and test tasks differ in goal specification.
\begin{itemize}
\item \textit{State and actions:} States are SE(3) object poses, estimated from wrist-camera scans using SAM3~\cite{carion2025sam} with depth reprojection and heuristic orientation estimation. Actions are 6-D joint-position commands plus a 1-D gripper command.
\item \textit{Object types:} Robot ($\mathtt{r}$), blocks ($\mathtt{b}$), box ($\mathtt{box}$).
\item \textit{Operators:} $\mathtt{Pick(?r, ?b)}$, $\mathtt{Insert(?r, ?b0, ?b1)}$.
\item \textit{Predicates:} $\mathtt{HandEmpty(?r)}$, $\mathtt{Holding(?r, ?b)}$, $\mathtt{On(?b0, ?b1)}$.
\item \textit{Hyperparameters:} Recovery skills follow OmniReset~\cite{yin2026omnireset}, with horizon $200$ and up to $5B$ environment steps. Concept learning uses $300$ augmented trajectories in Stage 1.
\end{itemize}

\item \textbf{Cornered Insertion-L.} \
This domain uses the same hardware setup but adds a lid and a drawer, requiring the robot to place the completed tower into the drawer. Although it shares the recovery skill with Cornered Insertion-S, its task structure leads to different discovered concepts and planning operators.
\begin{itemize}
\item \textit{State and actions:} Same as Cornered Insertion-S.
\item \textit{Object types:} Robot ($\mathtt{r}$), blocks ($\mathtt{b}$), box ($\mathtt{box}$), lid ($\mathtt{lid}$), drawer ($\mathtt{drawer}$).
\item \textit{Operators:} $\mathtt{Pick(?r, ?b)}$, $\mathtt{Insert(?r, ?b0, ?b1)}$, $\mathtt{RemoveLid(?r, lid)}$, $\mathtt{PickTower(?r, ?b0, ?b1)}$, $\mathtt{PlaceTower(?r, ?b0, ?b1, drawer)}$, $\mathtt{OpenDrawer(?r, drawer)}$, $\mathtt{CloseDrawer(?r, ?b0, ?b1, drawer)}$.
\item \textit{Predicates:} $\mathtt{HandEmpty(?r)}$, $\mathtt{Holding(?r, ?b)}$, $\mathtt{On(?b0, ?b1)}$, $\mathtt{LidOn(lid)}$, $\mathtt{LidOff(lid)}$, $\mathtt{DrawerOpen(drawer)}$, $\mathtt{DrawerClosed(drawer)}$.
\item \textit{Hyperparameters:} Same as Cornered Insertion-S.
\end{itemize}

\end{enumerate}

\section{Details about ReSYNC}\label{app:resync_details}

\myparagraph{PDDL Formulation:}
We use PDDL as the symbolic interface between learned neural abstractions and task-level planning.
Each typed object in the RF-MDP is declared as a PDDL object, each initial or learned predicate $\psi \in \Psi$ becomes a typed PDDL predicate, and each skill $\mathtt{C}\in\mathcal{C}$ is represented by a lifted PDDL operator whose parameters match the skill's relational signature.
An operator's preconditions specify the lifted predicates that must hold before executing the corresponding ground skill, while its add and delete effects describe the expected symbolic state change after the skill terminates.
At planning time, the current continuous state $\mathbf{x}$ is converted into an initial symbolic state by evaluating all predicate classifiers, and the task goal $g$ is provided as a set of goal predicates.
The planner searches over ground operators; during execution, each selected operator invokes its associated low-level skill policy, and the symbolic state is re-evaluated after skill termination to determine whether replanning is necessary.

\myparagraph{Details about Failure objects:}
In our experiments, failure objects are identified automatically from simulator interaction signals.
We first train an ``overfitting'' recovery skill on the $N$ user-provided failure states, so that executing the skill makes replanning successful from each of those states.
We then roll out this skill and mark as failure objects the objects that experience interaction forces above a fixed threshold.
When multiple failure objects are involved, we fit a pairwise Gaussian distribution between each failure object and the relevant object.
During both failure mining and relational data augmentation, we resample the states of all failure objects from their corresponding fitted Gaussians.

\begin{algorithm2e}[!t]
\caption{Predicate Selection}
\label{alg:predicate_selection}
\DontPrintSemicolon

\KwIn{Candidate predicates $\tilde{\Psi}^{\mathcal{N}}$, augmented dataset $\mathcal{D}^{\mathcal{N}}$, previous predicates $\Psi$}
\KwOut{Selected predicates $\Psi^{\mathcal{N}}$ and learned operators $\mathtt{Op}^{\mathcal{C}^\mathcal{N}}$}

Initialize $\Psi^{\mathcal{N}} \leftarrow \emptyset$, $J^* \leftarrow \infty$\;
\While{True}{
    $\psi^* \leftarrow \text{None}$, $J_{\min} \leftarrow \infty$\;

    \For{each $\psi \in \tilde{\Psi}^{\mathcal{N}} \setminus \Psi^{\mathcal{N}}$}{
        $\hat{\Psi} \leftarrow \Psi \cup \Psi^{\mathcal{N}} \cup \{\psi\}$\;
        Learn operators $\hat{\mathtt{Op}}^{\mathcal{C}^\mathcal{N}}$ from $\hat{\Psi}$ and $\mathcal{D}^{\mathcal{N}}$\;
        Evaluate planning score $J(\hat{\Psi}, \mathcal{D}^{\mathcal{N}})$\;

        \If{$J < J_{\min}$}{
            $J_{\min} \leftarrow J$, $\psi^* \leftarrow \psi$\;
        }
    }

    \If{$J_{\min} \ge J^*$}{
        \textbf{break}
    }

    $\Psi^{\mathcal{N}} \leftarrow \Psi^{\mathcal{N}} \cup \{\psi^*\}$\;
    $J^* \leftarrow J_{\min}$\;
}

Learn final operator set $\mathtt{Op}^{\mathcal{C}^\mathcal{N}}$ from $\Psi \cup \Psi^{\mathcal{N}}$\;
\Return $\Psi^{\mathcal{N}}, \mathtt{Op}^{\mathcal{C}^\mathcal{N}}$\;
\end{algorithm2e}

\myparagraph{Terminal Function Learning:}
At each training stage, we learn two terminal classifiers: one for the failed skill $\mathtt{C}^{\mathcal{F}}$ and one for the newly learned recovery skill $\mathtt{C}^{\mathcal{N}}$.
The terminal classifier $\beta_{\mathtt{C}^{\mathcal{F}}}$ is trained by labeling all states preceding the failure state $\mathbf{x}^{\mathcal{F}}$ as \texttt{False}, and the failure states $\mathbf{x}^{\mathcal{F}} \sim \mathcal{D}^{\mathcal{F}}$ as \texttt{True}.
Similarly, the terminal classifier $\beta_{\mathtt{C}^{\mathcal{N}}}$ is trained by labeling all states preceding the recovery goal state $\mathbf{x}'$ as \texttt{False}, and the goal states $\mathbf{x}'$ as \texttt{True}.
Both classifiers are implemented as neural networks and optimized using binary cross-entropy loss.
The terminal classifier $\beta_{\mathtt{C}^{\mathcal{F}}}$ is also used as the failure detector at test time for the RC baseline.

\myparagraph{Predicate Pool Learning:}
Following IVNTR~\cite{li2025IVNTR}, since $\mathcal{D}^{\mathcal{N}}$ is generated through skill execution, we can extract high-level transition tuples of the form $(\mathbf{x}, \underline{\mathtt{C}}, \mathbf{x}')$, where $\underline{\mathtt{C}}$ denotes a ground skill, and $\mathbf{x}, \mathbf{x}'$ denotes states prior and post the skill execution.
For a lifted predicate $\psi$, the lifted effect vector $\Delta^\psi$
induces a \emph{ground effect vector}
$\mathbf{t}^{\psi,\underline{\mathtt{C}}} \in \{-1,0,1\}^P$, where $P$ is the number of groundings of $\psi$ over the object set $\mathcal{O}$.
Each entry specifies whether a particular ground atom is expected to remain unchanged, be added, or be deleted by the execution of $\underline{\mathtt{C}}$.
Given a transition $(\mathbf{x}, \underline{\mathtt{C}}, \mathbf{x}')$,
we apply the predicate classifier to all groundings of $\psi$:
\begin{equation*}
\hat{\mathbf{v}}, \hat{\mathbf{v}}' =
\mathrm{Ground}(\mathbf{x}, \theta_\psi),\;
\mathrm{Ground}(\mathbf{x}', \theta_\psi)
\in [0,1]^P,
\end{equation*}
obtaining predicted truth values before and after the transition.
The effect vector $\mathbf{t}^{\psi,\underline{\mathtt{C}}}$ partitions ground atoms into
unaffected and affected subsets via masks
\begin{equation*}
\mathbf{m}_0 = \mathbb{I}\big(\mathbf{t}^{\psi,\underline{\mathtt{C}}} = 0\big),
\quad
\mathbf{m}^1 = \mathbb{I}\big(|\mathbf{t}^{\psi,\underline{\mathtt{C}}}| = 1\big),
\end{equation*}
with $S^1 = \{ p \mid m^1_p = 1 \}$ denoting affected atom indexes.

Predicate learning thus enforces consistency between the predicted predicate transitions
and the symbolic effects specified by $\Delta^\psi$.
For unaffected atoms, truth values should remain invariant, for affected atoms, transitions should match the specified add or delete semantics.
Accordingly, we define the per-transition loss
\begin{equation*}
\mathcal{L}(\mathbf{x}, \mathbf{x}', \theta_\psi)
=
\mathcal{L}_\mathrm{zero}
+
\mathcal{L}_\mathrm{one},
\end{equation*}
where
\begin{equation*}
\mathcal{L}_\mathrm{zero}
=
\mathrm{JS}\!\Big(
\hat{\mathbf{v}} \odot \mathbf{m}_0
\ \Big\|\
\hat{\mathbf{v}}' \odot \mathbf{m}_0
\Big)
\end{equation*}
penalizes violations of invariance, and
\begin{equation*}
\mathcal{L}_\mathrm{one}
=
\frac{1}{|S^1|}
\sum_{p \in S^1}
\mathrm{BCE}\Big(
[\hat{v}_p, \hat{v}'_p],
\big[\tfrac{1 - t_p}{2}, \tfrac{1 + t_p}{2}\big]
\Big)
\end{equation*}
penalizes incorrect add ($t_p=1$) or delete ($t_p=-1$) behavior.
Here, $\mathrm{JS}(\cdot\|\cdot)$ denotes Jensen–Shannon divergence and
$\mathrm{BCE}(\cdot,\cdot)$ denotes binary cross-entropy.
Aggregating across transitions and skills yields the dataset-level objective
\begin{equation*}
\mathcal{L}^{\mathcal{D}}(\theta_\psi)
=
\sum_{\mathtt{C} \in \mathcal{C}^\mathcal{N}}
\mathbb{E}_{(\mathbf{x},\underline{\mathtt{C}},\mathbf{x}') \sim \mathcal{D}_{\mathtt{C}}}
\mathcal{L}(\mathbf{x}, \mathbf{x}', \theta_\psi),
\end{equation*}
which can be optimized end-to-end using standard gradient-based methods,
without requiring explicit predicate labels.
After training with these labels, the classifier for $\psi$ is evaluated on a held-out validation set,
yielding a validation loss.
Predicates whose classifiers have validation loss below a fixed threshold are retained, forming the final discovered predicate set $\tilde{\Psi}^{\mathcal{N}}$.
To constrain the potential space of $\Delta^\psi$, we enumerate the effect profiles with nonzero entries only for a subset of skills, namely the recovery skill $\mathtt{C}^{\mathcal{N}}$, the failed skill $\mathtt{C}^{\mathcal{F}}$, and the skill following recovery.

\myparagraph{Predicate Selection:}
Motivated by previous work~\cite{silver2023predicateinvent}, \model{} explicitly selects a compact subset of predicates that are most useful for planning.
Given a set of predicate candidates $\tilde{\Psi}^{\mathcal{N}}$ discovered from augmented trajectories $\mathcal{D}^{\mathcal{N}}$, \model{} evaluates candidate predicate sets using a planning-driven objective $J(\hat{\Psi}, \mathcal{D}^{\mathcal{N}})$.
This objective measures the quality of the planning operators derived from $\hat{\Psi}$ along two dimensions: (i) whether the resulting operators can reproduce the skill sequences seen in $\mathcal{D}^{\mathcal{N}}$, and (ii) the efficiency of planning, measured by the number of node expansions required to find the high-level skill-plan.
Concretely, for a candidate predicate set $\hat{\Psi}$, we ground predicates over the states in $\mathcal{D}^{\mathcal{N}}$ and learn a corresponding operator set $\hat{\mathtt{Op}}^{\mathcal{C}'}$ following standard symbolic operator learning procedures~\cite{chitnis2021nsrt}.
We then evaluate $\hat{\mathtt{Op}}^{\mathcal{C}'}$ by planning over the augmented tasks and computing the resulting score $J$~\cite{silver2023predicateinvent}.
To identify an effective and compact abstraction, \model{} performs a hill-climbing search over
$\tilde{\Psi}^{\mathcal{N}}$.
At each iteration, the predicate whose inclusion yields the greatest improvement in the planning objective is added to the selected set.
The procedure terminates when no candidate provides further improvement or when a predefined threshold is reached.
The final output is the union of the current and the selected predicates $\Psi' = \Psi\cup\Psi^{\mathcal{N}}$ and the corresponding operator set
$\mathtt{Op}^{\mathcal{C}'}$, yielding the new abstract planning model.
For better understanding, we present an algorithm table for this process in \algoref{alg:predicate_selection}.

\myparagraph{Predicate Finetuning:}
After stage $(i+1)$, the skill set becomes $\mathcal{C}_{i+1} = \mathcal{C}_i \cup \{\mathtt{C}^{\mathcal{N}}_{i+1}\}$.
For each existing predicate $\psi \in \Psi_i$, we extend its effect vector $\Delta_i^\psi \in \{-1,0,1\}^M$ to $\Delta_{i+1}^\psi \in \{-1,0,1\}^{M+1}$ by computing the effect of the new operators using previous operator learning approach~\cite{chitnis2021nsrt}.
We then finetune the previously learned predicate classifiers using the updated effect vectors $\Delta_{i+1}^\psi$ on the newly augmented trajectories $\mathcal{D}^{\mathcal{N}}_{i+1}$, following the effect-consistent learning objective introduced above.


\section{Invariance to Curriculum Order}\label{app:revsed_order}
\begin{table}[htbp]
    \tablestyle{2pt}{0.8}
    \scriptsize
    \fontsize{8}{10}\selectfont
    \caption{Robustness to curriculum ordering in \textit{Blocked Stacking}. Reversing the default task sequence ($\mathcal{T}_2 \to \mathcal{T}_1$) yields performance comparable to the original ordering, indicating robustness to this change in curriculum order.}
    \label{tab:curriculum}
    \begin{center}
        \begin{tabular}{c|cc|cc|cc}
        \toprule[1.5pt]
        Task  & \multicolumn{2}{c|}{In} & \multicolumn{2}{c|}{Old} & \multicolumn{2}{c}{New} \\
        Metric & Acc   & Var   & Acc   & Var   & Acc   & Var \\
        \midrule
        Default & 0.64  & \multirow{2}[2]{*}{-6.0\%} & 0.78  & \multirow{2}[2]{*}{1.0\%} & 0.83  & \multirow{2}[2]{*}{-6.0\%} \\
        Reversed & 0.58  &       & 0.79  &       & 0.77  &  \\
        \bottomrule[1.5pt]
        \end{tabular}%
    \end{center}
\end{table}

We further assess the sensitivity of \model{} to curriculum design by reversing the sequence of failures ($\mathcal{T}_2 \to \mathcal{T}_1$) in the \textit{Blocked Stacking} domain.
As summarized in \tref{tab:curriculum}, the robot achieves comparable success rates on the final test tasks regardless of the training order.
This experiment suggests that \model{} can integrate the learned relational concepts under either of these two curriculum orders.

\section{Test-time Planning Efficiency}\label{app:plan_eff}
\begin{figure}[htbp]
  \begin{center}
    \includegraphics[width=0.55\linewidth]{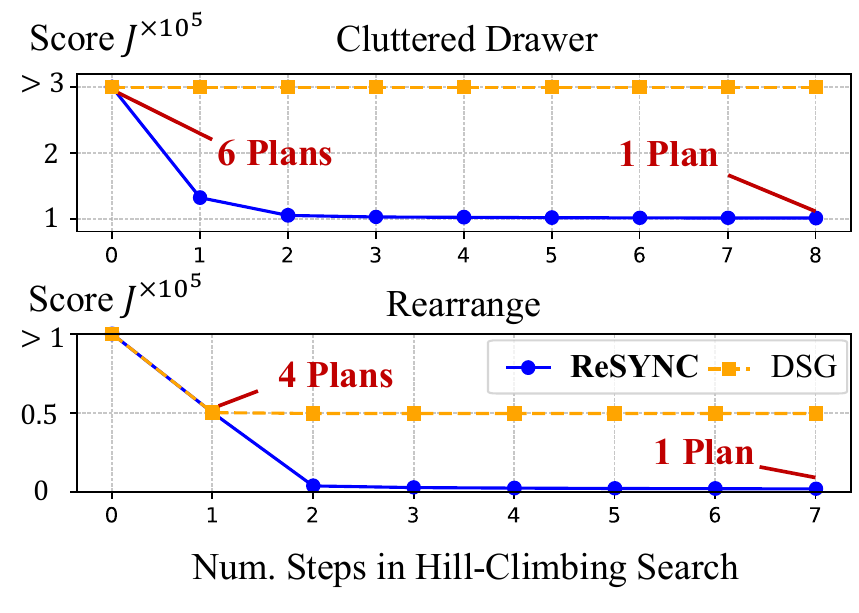}
    \caption{Planning efficiency comparison. \model{} learns compact relational concepts that minimize the planning objective and yield single-plan execution, while DSG-M~\cite{bagaria2025im-dsg} relies solely on terminal predicates, resulting in inefficient planning.}
	\label{fig:objective}
  \end{center}
\end{figure}

\model{} achieves substantially higher planning efficiency than prior work such as DSG-M~\cite{bagaria2025im-dsg}.
As shown in \fref{fig:objective}, DSG-M learns predicates that primarily characterize terminal outcomes of individual skills.
Although such predicates can be useful for specifying local skill order, they provide limited guidance for abstract planning: many potential paths could lead to the goal state.
As a result, the planner must explore a larger search space, leading to high planning objectives and multiple redundant high-level plans.

In contrast, \model{} explicitly selects predicates according to their impact on the planning objective.
The discovered predicates therefore capture not only whether a skill can terminate successfully, but also their contribution to the test-time planning efficiency~\cite{silver2023predicateinvent}.
This allows \model{} to prune spurious symbolic transitions and retain a compact set of meaningful planning choices.
Consequently, at test time, \model{} consistently produces a single reliable high-level plan, while DSG-M often produces several redundant alternatives that increase planning cost and reduce execution efficiency.

\section{Noisy Initial Knowledge}\label{app:imperfect}

\begin{table}[htbp]
    \tablestyle{3pt}{0.8}
    \scriptsize
    \fontsize{8}{12}\selectfont
    \caption{Robustness of \model{} under noisy initial knowledge. Adding noise to the initial predicate classifiers in Blocked Stacking does not significantly reduce the performance of \model{}.}
    \label{tab:noisy-predicates}
        \begin{center}
        \begin{tabular}{ccccccc}
        \toprule[1.5pt]
        Noise STD & 0     & 0     & 0.16  & 0.16  & 0.38  & 0.38 \\
        \midrule
        Task  & $\mathcal{T}^\mathrm{train}_1$    & $\mathcal{T}^\mathrm{test}_1$     & $\mathcal{T}^\mathrm{train}_1$    & $\mathcal{T}^\mathrm{test}_1$     & $\mathcal{T}^\mathrm{train}_1$    & $\mathcal{T}^\mathrm{test}_1$\\
        Success & 0.73  & 0.76  & 0.70  & 0.69  & 0.66  & 0.70 \\
        \bottomrule[1.5pt]
        \end{tabular}%
    \end{center}
\end{table}

While improving the initial knowledge, $\{\mathcal{C}_0, \Psi_0,\mathtt{Op}^{\mathcal{C}_0}\}$, is complementary to our work, we find that ReSYNC is robust to imperfect initialization.
In \tref{tab:noisy-predicates}, we introduced 16\%$\sim$38\% perceptual noise (w.r.t the mean decision boundary) into the initial predicates.
ReSYNC was able to discard unreliable predicates and replace them with newly learned ones, resulting in robust test-time planning performance.

\section{Representative Failures in Real-World Experiments}\label{app:failure}

\begin{figure*}[!t]
	\centering
	\includegraphics[width=\textwidth]{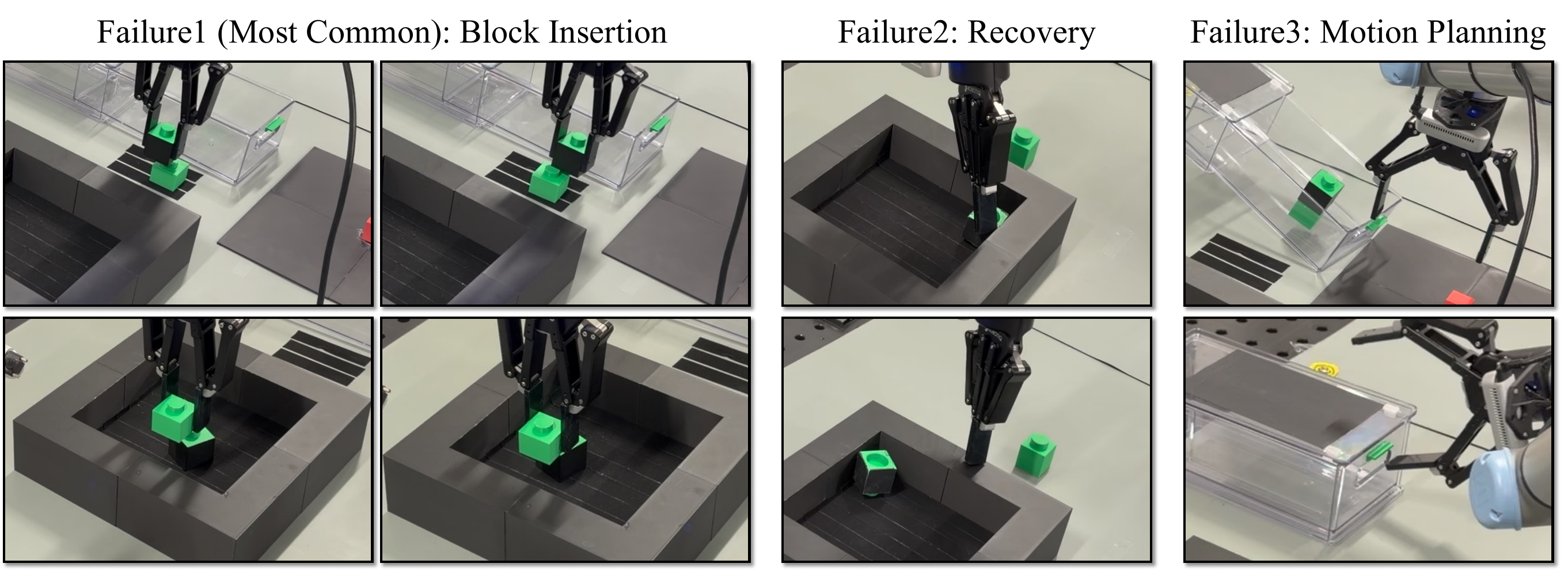}
    \vspace{-1em}
	\caption{Representative failures in our real-robot experiments.}
	\label{fig:failures}
\end{figure*}

As shown in \fref{fig:failures}, we present representative real-robot failure cases.
Overall, task planning is reliable, benefiting from the discovered neural predicates and the planning-driven learning objective.
The most common failures occur during insertion, where perception noise can misalign the placement pose.
This failure mode arises because insertion is executed by a scripted Gaussian-filter-based controller rather than a learned policy, and our system does not use force feedback.
The learned RL recovery skill can also fail due to the sim-to-real gap.
For example, in the top row, the gripper becomes stuck in the gap, leading to an unrecoverable state; in the bottom row, misaligned image observations cause the gripper to collide with the box.
Finally, high-precision skills such as drawer pulling and pushing remain sensitive to perception and control noise.
These observations motivate future research on long-horizon planning with dexterous, high-precision, and contact-rich skills, moving beyond primarily prehensile pick-and-place skills~\cite{li2025IVNTR}, where many open problems remain.

\section{Extended Discussions on the Limitations}\label{app:assumptions}
Below, we discuss potential approaches for relaxing each assumption in \model{}.

\myparagraph{User-provided learning curriculum}: One potential approach to automate curriculum design is to allow a coding agent to explore and understand the simulated environment~\cite{fu2026cap}.
Specifically, we can prompt a coding agent to attempt planning in various tasks and understand the different failures.
Then the coding agent could automate the process of failure-recovery MDP creation and policy learning.

\myparagraph{The assumption of being able to identify failure states}: This is a common assumption in previous work that learning to recover from failures~\cite{vats2024recoverychaining}.
There is also orthogonal work on failure detection with foundation models~\cite{duanaha,llm3,xu2025can} that can be combined with our work to enable fully autonomous identification of failure states and subsequent recovery. A preliminary visual pipeline for identifying the failure-causing object is evaluated in \appref{app:visual_failure_object}.

\myparagraph{The assumption of being able to identify the single relevant object}:
We follow previous work~\cite{silver2022skills} to derive the failure-relevant object from the failed ground skill.
In a more general case where multiple relevant objects are involved, multiple pairwise Gaussian distributions need to be fitted.

\myparagraph{Gaussian state distributions}:
In \model{}, the sampled states from the pairwise Gaussians are validated by resetting the physics simulator and stepping it forward for 20 steps.
We retain only those states that remain stable under simulation.
A promising direction to relax this assumption is to leverage recent diffusion-based generative models for state sampling~\cite{yang2025skillwrapper,shah2024reals,yang2023diffusion}.
For example, we could first learn neural predicates from real trajectory data, and then perform classifier-guided diffusion~\cite{luo2025concept} to generate valid and diverse states for the relational data augmentation phase.
This approach would allow us to move beyond simple Gaussian distributions and capture more complex state distributions.
We also evaluate a preliminary rejection-sampling alternative in \appref{app:rejection_sampling}.

\myparagraph{Partial observability from state projection}:
Projecting the state onto the failure and relevant objects omits features of other objects. The projected observation may therefore be insufficient to predict skill outcomes, even when the full environment state is Markovian. This can cause recovery failures; learning which additional context to include is future work.

\myparagraph{Simplifying symbolic learning}:
Much of the framework's complexity arises from learning symbolic abstractions. We explored whether LLM agents could infer these abstractions directly. Preliminary observations indicate that low-level RL skills without clear semantics remain difficult for these agents to interpret. A quantitative evaluation of this alternative remains future work.

\myparagraph{Scalability of compositional relational augmentation}:
ReSYNC uses neural skill policies and predicate classifiers in batched environments, enabling parallel GPU execution. In Rearrange, with six objects, augmentation takes approximately two hours on a single A100 GPU.
Learning predicates from scratch minimizes prior assumptions but increases computational cost as the object set grows.
Pre-trained foundation models could initialize~\cite{wang2025unipred} or guide~\cite{athalye2024predicate} predicate learning, reducing the amount of augmented data needed to cover object combinations.

\clearpage
\section{Additional Experiments}\label{app:additional_experiments}
We present three additional experiments that examine the state-sampling assumption, contact-rich skill composition, and identification of failure-causing objects. These extend the evaluations in \sref{sec:experiments}; the sampling and visual-identification studies are preliminary alternatives to components of the main pipeline.

\subsection{Relational Augmentation with Rejection Sampling}\label{app:rejection_sampling}
\myparagraph{Motivation and procedure:}
Our default augmentation procedure fits pairwise Gaussians to states before and after a recovery skill. To investigate whether Gaussian sampling is essential, we evaluate a rejection-sampling alternative in Icy Transport and Cluttered Drawer.
We first learn neural predicates from recovery trajectories and extract the new skill's effect set. We then sample initial states from $P(\mathbf{x}_0)$ and retain only those whose grounded predicate values match the effect annotations observed immediately before or after the new skill's execution.
This supplies additional states for relational augmentation without fitting the pairwise Gaussian distributions used by the default augmentation procedure.

\begin{table}[htbp]
    \centering
    \small
    \setlength{\tabcolsep}{9pt}
    \begin{tabular}{l|cc|cc}
    \toprule
    & \multicolumn{2}{c|}{Icy Transport} & \multicolumn{2}{c}{Cluttered Drawer} \\
    Method & $\mathcal{T}^{\mathrm{train}}$ & $\mathcal{T}^{\mathrm{test}}$ & $\mathcal{T}^{\mathrm{train}}$ & $\mathcal{T}^{\mathrm{test}}$ \\
    \midrule
    RC+IVNTR & 0.50 & 0.44 & 0.69 & 0.59 \\
    + Rejection sampling & \textbf{0.86} & \textbf{0.76} & \textbf{0.74} & \textbf{0.73} \\
    \bottomrule
    \end{tabular}
    \caption{Preliminary success rates with rejection-based relational augmentation. The RC+IVNTR comparison uses recovery trajectories without relational augmentation.}
    \label{tab:rejection_sampling}
\end{table}

\myparagraph{Results and limitations:}
As shown in \tref{tab:rejection_sampling}, rejection sampling improves held-out success over RC+IVNTR by 32 percentage points in Icy Transport and 14 percentage points in Cluttered Drawer. These results suggest that the benefit of relational augmentation is not specific to Gaussian sampling.
However, the preliminary rejection procedure is slow and retains noisy states. The experiment addresses the augmentation sampler; Gaussian sampling in failure mining is outside the scope of this comparison.

\clearpage
\subsection{Contact-Rich Flipped Stacking}\label{app:flipped_stacking}
\myparagraph{Task and generalization:}
Flipped Stacking requires contact-rich manipulation to reorient blocks before stacking them. Training tasks involve a two-level tower with the top block initially flipped. From recovery experience, \model{} learns a \texttt{Flip} skill and predicates that allow the skill to be incorporated into abstract planning.
Held-out tasks require a three-level tower with different top/bottom blocks flipped, testing reuse of the learned skill with new object bindings and longer compositions.

\begin{figure}[htbp]
    \centering
    \includegraphics[width=\linewidth]{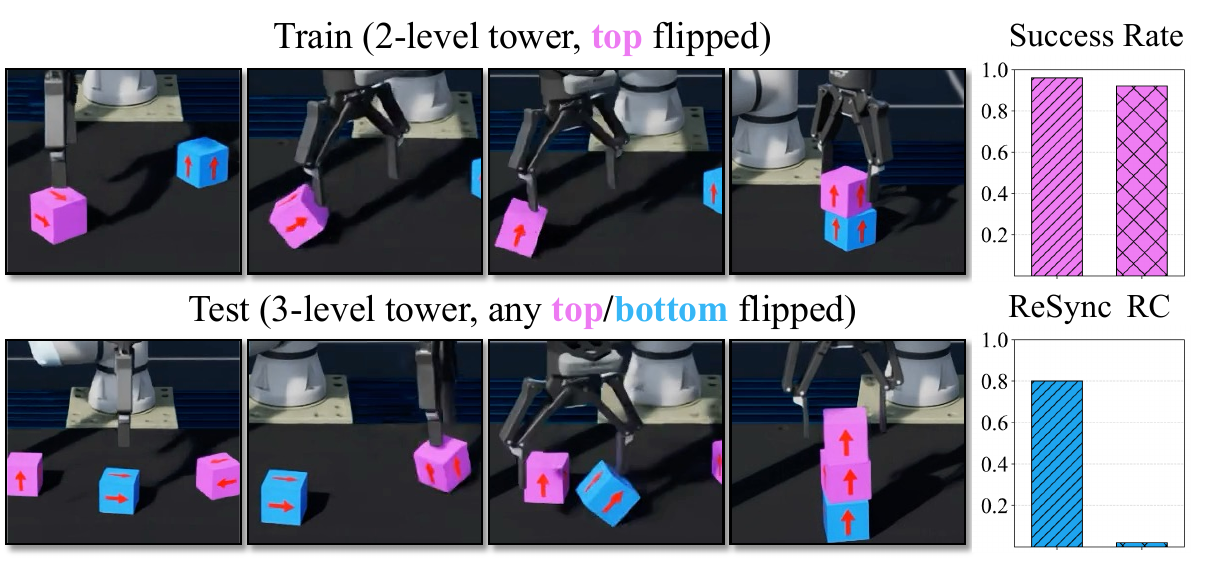}
    \caption{Flipped Stacking in simulation. Top: training on two-level towers with the top block flipped. Bottom: generalization to three-level towers with different top/bottom blocks flipped. Both methods succeed frequently in the training setting, while \model{} generalizes substantially better than Recovery Chaining (RC).}
    \label{fig:flipped_stacking}
\end{figure}

\myparagraph{Results:}
The success-rate plots in \fref{fig:flipped_stacking} show approximately 96\% training success for \model{} and 92\% for RC. On held-out tasks, \model{} achieves approximately 80\% success, while RC achieves approximately 2\%.
These results extend the evidence for contact-rich skill composition: learning to recover in one stacking context provides reusable skills and predicates for planning in new contexts.
This evaluation is in simulation; sim-to-real transfer of the \texttt{Flip} skill remains future work.

\subsection{Visual Identification of Failure-Causing Objects}\label{app:visual_failure_object}
The main simulation pipeline identifies failure-causing objects using interaction signals (\appref{app:resync_details}). We additionally test an image-based alternative in Cluttered Drawer: SAM3~\cite{carion2025sam} segments objects from images, and GPT-5.6 identifies the object responsible for the failure.
The pipeline correctly identifies the responsible object in 89 of 100 evaluated states (89\% accuracy).
This preliminary result suggests a route toward identifying failure-causing objects from visual observations. End-to-end recovery and planning with imperfect visual object identification remain future work.

\end{document}